\documentclass[11pt]{article}
\PassOptionsToPackage{table}{xcolor}
\usepackage{acl}
\usepackage{times}
\usepackage{latexsym}
\usepackage[T1]{fontenc}
\usepackage[utf8]{inputenc}
\usepackage{inconsolata}
\usepackage{microtype}
\usepackage{graphicx}
\usepackage{algorithm}
\usepackage{algpseudocode}
\usepackage{enumitem}
\usepackage{booktabs}
\usepackage{amsmath}
\usepackage{tcolorbox}
\usepackage{subcaption}
\usepackage{multirow}

\definecolor{beigebackground}{RGB}{245, 245, 220}
\definecolor{greenText}{HTML}{2ECC71}
\definecolor{redText}{HTML}{E74C3C}

\newcommand{\idtt}[1]{{\renewcommand{\UrlFont}{\ttfamily}\nolinkurl{#1}}}

\def\website{\url{https://ai-mh.github.io/SQPsych.html}}
\DeclareRobustCommand{\pipeline}{{\small\textsf{SQPsych}}}

\DeclareRobustCommand{\llm}[1]{{\small\textsf{SQPsychLLM}\textsubscript{#1}}}

\DeclareRobustCommand{\conv}[2]{{\small\textsf{SQPsychConv}\textsubscript{#1}\textsuperscript{\texttt{#2}}}}

\tcbuselibrary{skins, breakable}

\title{Roleplaying with Structure: Synthetic Therapist-Client Conversation Generation from Questionnaires}

\author{
  \textbf{Doan Nam Long Vu}\textsuperscript{1,\textdagger}, \textbf{Rui Tan}\textsuperscript{2},
  \textbf{Svenja Jule Francke}\textsuperscript{2}, \textbf{Lena Moench}\textsuperscript{3},
  \textbf{Daniel Woiwod}\textsuperscript{2},\\\textbf{Florian Thomas-Odenthal}\textsuperscript{2},
  \textbf{Sanna Stroth}\textsuperscript{2}, \textbf{Tilo Kircher}\textsuperscript{2},
  \textbf{Christiane Hermann}\textsuperscript{3},\\\textbf{Udo Dannlowski}\textsuperscript{4},
  \textbf{Hamidreza Jamalabadi}\textsuperscript{2}, \textbf{Simone Balloccu}\textsuperscript{1,\textdagger},
  \textbf{Shaoxiong Ji}\textsuperscript{1,5,6}\thanks{Work done while with TU Darmstadt}\\
  \textsuperscript{\textdagger}Natural Language Processing for Expert Domains (ExpNLP),\\
  \textsuperscript{1}Technical University of Darmstadt,
  \textsuperscript{2}Philipps-University Marburg,\\
  \textsuperscript{3}Justus Liebig University Giessen,
  \textsuperscript{4}University of Münster,\\
  \textsuperscript{5}ELLIS Institute Finland,
  \textsuperscript{6}University of Turku}

\begin{document}

\maketitle

\begin{abstract}
Large Language Models (LLMs) are promising tools for synthetic data generation in mental health. However, privacy policies and restrictions forced previous work to rely mainly on generic information. We present a comprehensive corpus of synthetic therapist-client conversations generated through LLMs. We construct our generation pipeline,~\pipeline~(Structured Questionnaire-based Psychotherapy), which uses real structured client profiles and psychological questionnaires without leaking any sensitive data. We fine-tune various open-weight LLMs on our generated corpus,~\conv{}{}, and test them through both automatic benchmarks and human evaluation with trained psychotherapists. We find that standard benchmarks do not adequately capture the strengths of our dataset, but expert judgment shows that~\pipeline~makes LLMs significantly better at therapist roleplaying. Experts also consistently prefer therapy sessions generated by our models compared to other mental-health-oriented LLMs. We release our code, fine-tuned models~\llm{}, and corpora at~\website.
\end{abstract}

\section{Introduction}

\begin{figure}[t]
    \centering
    \includegraphics[width=0.96\linewidth]{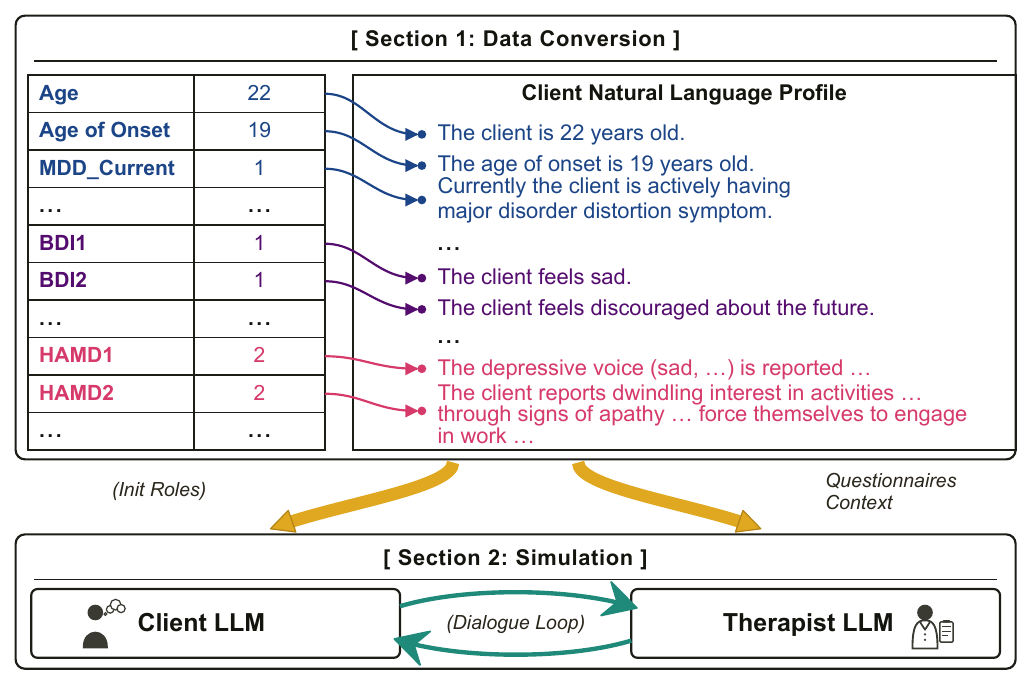}
    \caption{Our pipeline begins by taking numerical client data and questionnaire results and converting them into plain English in Step 1. To illustrate this, we provide an artificial example of how structured metadata, such as Age, Age of Onset, and MDD\_Current and scores like BDI1, BDI2, HAMD1, and HAMD2 are mapped to their descriptive definitions. This structured-to-natural-language conversion turns raw numbers into a narrative format, which then serves as the foundation for Step 2, where LLMs use that text to simulate realistic patient conversations.}
    \label{fig:s-label}
\end{figure}

Acquiring therapy-client dialogues from clinical institutions is challenging due to strict privacy laws such as HIPAA (US), GDPR (EU) and PIPL (China), along with ethical issues regarding client confidentiality for further usage~\cite{de2022chatbots}. These regulations restrict access to large and varied datasets necessary to build AI systems for mental health care. Furthermore, historically, clinical psychology interviews and counseling sessions were not routinely recorded or transcribed. In practice, questionnaires were often derived from insights distilled informally during these sessions. With the advent of LLMs, there is now potential to leverage these structured processes and data sources to synthetically generate more meaningful outcomes, such as therapy-client dialogues. 

Synthetic therapy-client conversations offer valuable opportunities in mental health care and clinical research~\cite{sharma2021towards}. However, previous works have not investigated generating synthetic data with real-world client information~\cite{lee-etal-2024-cactus, qiu2023smile} due to privacy-compliance concerns. While privacy laws prevent sharing real data, the lack of grounded synthetic data has prevented the community from validating whether LLMs can truly simulate specific therapeutic skills. In this paper, we conduct the first study to investigate these differences. 

We introduce an LLM-based generation pipeline that produces synthetic counseling dialogues conditioned on structured client data, standardized mental health assessments, and psychologically informed guidance. These conversations are tailored to specific mental health conditions (e.g., major depressive disorder and anxiety). By integrating psychological expertise and best practices (specifically Cognitive Behavioral Therapy - CBT)~\cite{beck1963thinking}, and bridging structured psychological data with natural language conversation, our approach facilitates the generation of clinically relevant dialogues. Figure~\ref{fig:s-label} illustrates our main concept. Our contributions are as follows:
\begin{itemize}[topsep=0pt, partopsep=0pt, parsep=0pt, itemsep=0pt, leftmargin=*]
\item We introduce~\pipeline~(\textbf{S}tructured \textbf{Q}uestionnaire-based \textbf{Psych}-
otherapy), a pipeline that transforms structured client metadata and questionnaires into natural language text and employs role-playing between therapist and client personas to generate counseling conversations.
\item 
We employ seven open models (23B--123B) for local synthetic data generation, conditioned on real-world structured client information and questionnaires. We then leverage our synthetic dataset ~\conv{}{},
distilled from the larger LLMs, to fine-tune seven smaller models.

\item 
We perform a holistic evaluation with human experts, automatic metrics, and mental health benchmarks. Our models consistently beat previous finetuned SOTA, but vanilla LLMs lead with better performance. Since benchmarks only assess multiple-choice questions, we conduct a separate evaluation with human experts. Our models are consistently preferred for their empathy and correct capturing of therapy dynamics.

\item
We perform an ablation study, generating an additional corpus through~\pipeline~conditioned on client metadata only, without clinical questionnaires.
In a blind expert evaluation, we find enhanced therapist and client role-playing. 

\item We release both corpora to help the research community understand the impact of clinical grounding in synthetic data generation.

\end{itemize}

\section{Related Work} \label{sec:related} 
We present the recent related works on the use/creation of synthetic data related to mental health task and counseling conversation.

\subsection{Synthetic Data in Mental Research}
LLMs have been increasingly employed to generate synthetic data in mental health research. \citet{mori-etal-2024-towards} utilized GPT-3 to create the HeadRoom dataset and investigated demographic variations in depression-related content, revealing that while LLMs can reproduce real-world stressor patterns, they may also reflect biases. \citet{sun-etal-2024-collab} proposed MSIC, a system for synthesizing multi-visit electronic health records (EHR) that ensures medical plausibility across visits. Furthermore, \citet{ghanadian-etal-2025-improving} leveraged LLMs guided by social context factors to generate synthetic suicidal ideation, achieving high performance with synthetic data alone, and progressively better results when combined with 30\% real data.

\subsection{Synthetic Counseling Dataset}
Synthetic data produced by LLMs is also frequently used to refine downstream models intended for mental health counseling. These datasets are typically distilled from large, high-performance models, particularly the GPT family from OpenAI. \citet{cabrera-lozoya-etal-2025-synthetic} used GPT-3 to generate therapy-style dialogues, incorporating human-in-the-loop annotation to facilitate empathy detection in Reddit counseling contexts. \citet{qiu2023smile} started with existing public single-turn Question-Answer pairs (e.g., from mental health forums) and prompted \texttt{gpt-3.5-turbo} to expand each single-turn interaction into a coherent multi-turn conversation dataset (SMILE) in Chinese. Most recently, \citet{lee-etal-2024-cactus} employed \texttt{gpt-4o} within a goal-oriented and structured CBT framework to generate an extensive synthetic dataset CACTUS for mental health counseling.

Crucially, these previous works (e.g., CACTUS, SMILE) on synthetic mental health counseling \textbf{relied exclusively on proprietary, API-based models (e.g., GPT)}. However, such approaches are often incompatible with strict real-world data privacy regulations (e.g., in Europe) that prohibit transmitting sensitive clinical data to external infrastructure. Consequently, these methods are infeasible in settings that require strict compliance. Our work utilizes local models to address these critical barriers. This methodology ensures regulatory compliance, data privacy, transparency, and full control over the data generation pipeline.

\section{Methods}
\label{sec:method}
We propose \textbf{\pipeline}, a pipeline that converts structured mental health data into synthetic therapist-client dialogues. This approach integrates standardized client metadata and questionnaires with targeted role-based generation strategies to produce clinically grounded conversations. We describe the core components of our methodology in this section; detailed information regarding the full generation workflow is provided in Appendix~\ref{app:pipline-details}.

\subsection{Structured Data Conditioning \& Conversion}
\label{sec:data_inputs}
Our pipeline is conditioned on structured information comprising \textbf{general client demographics} (e.g., age, gender) and \textbf{standardized questionnaire results} (e.g., PHQ-9, BDI). Much of this information is numerical or categorical in nature, so we convert it into natural language for coherent text generation. Following~\citet{hegselmann2023tabllm}, we serialize each variable as a JSON object, with each key mapped to a variable name, and each value mapped to its textual description.

\subsection{Dual-role Generation Strategy}
\label{sec:conversation_generation}
Previous approaches, such as CACTUS~\cite{lee-etal-2024-cactus} and SMILE~\citep{qiu2023smile}, rely on a single model to simulate both roles simultaneously. In contrast, we employ a dual-agent framework where separate models independently represent the therapist and client, each operating with distinct reasoning processes and information access. This separation enhances dialogue dynamics, as each agent maintains a unique perspective and responds based on role-specific knowledge and objectives~\cite{wang-etal-2024-notechat}.

Both the client and therapist agents are conditioned on the client's background information and questionnaire data. The client agent generates responses that reflect the client's psychological state and personal circumstances, expressing \emph{emotions and hesitation}, grounded in personal data and questionnaire-derived insights. Conversely, the therapist agent operates according to therapeutic guidelines and intervention strategies and applies CBT skills (e.g., \emph{paraphrasing, validation}).

\subsection{Turn-Based Generation Workflow}
\label{sec:generation_workflow}
The conversation unfolds via a four-step role-playing therapeutic simulation.
\begin{enumerate}[topsep=0pt, partopsep=0pt, parsep=0pt, itemsep=0pt, leftmargin=*]
    \item \textbf{Initialization:} We instantiate two distinct LLMs, one as the therapist and the other as the client. 
    \item \textbf{Turn-by-turn interaction:} the two models iteratively generate an utterance, with the therapist one starting from an initial utterance. As this exchange continues, both models incorporate the full conversation history at each step to maintain contextual consistency. 
    \item \textbf{Termination criterion:} The dialogue progresses until the therapist model issues an explicit signal acknowledging the session's conclusion via an \texttt{[/END]} token. To ensure sufficient conversational depth, a minimum of 15 turns is required before termination is permitted~\citep{lee-etal-2024-cactus}.
    \item \textbf{Post-processing:} Following generation, we post-process the dialogue using regular expressions to remove redundant artifacts or internal monologues. We do this to filter out undesired outputs like `\textit{thinking}' tokens prior to the actual utterance.
\end{enumerate}

\section{Experiments}
\label{sec:experiments}
\subsection{Dataset \& Structured Input}

We utilize the questionnaire data and client demographics collected by \citet{kircher2019-db}, which are available upon request. This corpus comprises 2,090 anonymized clients, categorized into a \textbf{control} group and an \textbf{MDD} (Major Depressive Disorder) group. The primary distinction between these groups lies in their cognitive patterns. Specifically, cognitive challenges in clients with MDD persist significantly longer than those in the control group. To the best of our knowledge, this is the first work to use this dataset with LLMs.

\begin{table}[t]
\centering
\resizebox{\linewidth}{!}{
\begin{tabular}{@{}lll@{}}
\toprule
\textbf{Abb.}       &\textbf{Checkpoint}                               & \textbf{\# Par.} \\ \midrule
\texttt{mistral}    &\url{mistralai/Mistral-Large-Instruct-2407}        & 123B \\
\texttt{command}    &\url{CohereLabs/c4ai-command-a-03-2025}            & 111B\\
\texttt{qwen2.5}    &\url{Qwen/Qwen2.5-72B-Instruct}                     & 72B   \\
\texttt{llama3.3}   &\url{meta-llama/Llama-3.3-70B-Instruct}            & 70B  \\
\texttt{nemotron}   &\url{nvidia/Llama-3_3-Nemotron-Super-49B-v1}       & 49B \\
\texttt{qwq}        &\url{Qwen/QwQ-32B}                                 & 32B\\ 
\texttt{gemma}      &\url{google/gemma-3-27b-it}                        & 27B \\\bottomrule
\end{tabular}
}
\caption{Model checkpoints from Huggingface that we use in our paper to generate synthetic therapeutic conversations.}
\label{tab:open-model}
\end{table}

We incorporate general client information, including demographic attributes (\emph{age}, \emph{gender}, \emph{education}, and \emph{employment}) alongside family background metrics such as \emph{genetic risk}, \emph{parental education}, and \emph{number of children}. Additionally, \textbf{clinical characteristics} related to MDD, such as \emph{age of onset} and \emph{episode duration}, are included. We specifically select \textbf{questionnaire results} from established MDD assessment tools, including the \emph{Hamilton Depression Rating Scale} (HAM-D)~\cite{hamilton1980ratingdp}, the \emph{Hamilton Anxiety Rating Scale} (HAM-A)~\cite{hamlton1959theassessment}, and the \emph{Beck Depression Inventory} (BDI)~\cite{beck1974depression}. As these questionnaire datasets are originally in tabular format, we convert them into natural language using pre-defined descriptions of the questionnaires themselves. A detailed example of this conversion is provided in Appendix~\ref{app:numerical_to_text}.

\subsection{Synthetic Datasets Generation}
\label{sec:datasets}

\begin{table}[t]
\centering
\resizebox{\linewidth}{!}{
\begin{tabular}{@{}lrrr@{}}
\toprule

\textbf{Dataset}     & \textbf{\# Utt.} & \textbf{\# Avg. turns} & \textbf{\# Tok./utt.} \\\midrule
\multicolumn{4}{l}{\textit{Previous Works}} \\
\midrule
CACTUS~\cite{lee-etal-2024-cactus}  & 995,512      & 15.263       & 27.051                   \\
Psych8k~\cite{liu2023chatcounselor} & 16,374       & 1            & 54.685                   \\
\midrule
\multicolumn{4}{l}{\textit{Our work}} \\
\midrule
\conv{command}{}     & 64,760       & 17.451       & 51.019                   \\
\conv{gemma}{}       & 71,000       & 16.999       & 51.790                   \\
\conv{nemotron}{}    & 64,238       & 15.911       & 51.432                   \\
\conv{mistral}{}     & 98,342       & 23.119       & 31.098                   \\
\conv{llama3.3}{}    & 101,694      & 24.599       & 32.627                   \\
\conv{qwen2.5}{}     & 64,488       & 15.534       & 34.489                   \\
\conv{qwq}{}         & 77,134       & 18.601       & 26.291                   \\ \bottomrule
\end{tabular}
}
\caption{Dataset statistics comparing our approach to previous works on mental health counseling. On average, our datasets have more tokens/utterance than CACTUS, and are close to Psych8k.}
\label{tab:data-stat}
\end{table}

Although the clinical questionnaire dataset from~\citet{kircher2019-db} contains no Personally Identifiable Information (PII), its access terms impose strict requirements. Specifically, data privacy regulations restrict the usage of this clinical data to controlled, audited environments. This constraint prohibits transmitting the data to third-party services such as OpenAI or Azure. Consequently, we are restricted to employing open models to guarantee full data protection and regulatory compliance. We note that this also lets us actively contribute to open language modeling, rather than relying on proprietary vendor LLMs.

Table~\ref{tab:open-model} lists the seven open models utilized in our study. We employ vLLM~\cite{kwon2023efficient} to host these models locally and manage inference requests. The models are deployed on four A100 GPUs, each equipped with 80GB of VRAM, utilizing BF16 precision. We denote our synthetically generated datasets as~\conv{}{}; for instance,~\conv{mistral}{} refers to the dataset generated by the \texttt{mistral} model. Table~\ref{tab:data-stat} presents the dataset statistics, including the total number of utterances, average turns per dialogue, and average tokens per utterance. Generating a typical 15-turn conversation takes approximately three minutes. In total, we generated 2090 conversations per model, in $\approx$ 4.5 days.

\subsection{Baselines and Model Finetuning}
\label{sec:baseline}
After finishing the generation of our synthetic conversation dataset, we fine-tune Llama3-8B-Instruct~\cite{grattafiori2024llama} on each portion of the synthetic dataset to simplify comparison with baseline models which are also used to create their own domain-specific models. We use \llm{MODEL} for Llama3-8B-Instruct finetuned on the corpus produced by a specific LLM in Section~\ref{sec:datasets} (e.g.,~\llm{mistral} corresponds to LLaMA-8B-Instruct, fine-tuned on dialogues produced by \texttt{mistral},~\conv{}{mistral}).

We compare our fine-tuned models against established baselines derived from real data or proprietary models, to assess the transfer of counseling skills. We select baselines specifically designed for counseling and mental health applications. First, we utilize MentaLLaMA~\cite{yang2023mentalllama}, the first open-source LLM built upon Llama-2-7b-chat-hf~\cite{touvron2023llama}, which is optimized to detect and explain mental health signals in social network text.
Second, we employ CAMEL, which is based on Llama3-8B-Instruct. CAMEL is trained on the CACTUS dataset~\cite{lee-etal-2024-cactus}, a multi-turn synthetic dialogue corpus generated by \texttt{gpt-4o} using a structured CBT framework to mimic realistic counseling sessions.
Finally, to incorporate a baseline derived from real-world data, we utilize the Psych8k dataset (which contains real interactions rewritten by \texttt{gpt-4}). 
As this data is available in an instruction-tuning format, we perform a full BF16 fine-tuning of Llama3-8B-Instruct on Psych8k, using the same hyperparameters as CAMEL (5 epochs, learning rate $2\text{e-}4$).
We use the same hyperparameters for finetuning~\llm{}.

\section{Evaluation}

\subsection{Automatic Evaluation} 
We first evaluate our approach using three distinct benchmarks: CounselingBench~\cite{nguyen-etal-2025-large}, CBT Bench~\cite{zhang-etal-2025-cbt}, and an LLM Panel~\cite{verga2024replacing} (composed of multiple LLM-based judges). These assessments measure the performance of models fine-tuned on our synthetic dataset.

\subsubsection{CounselingBench} 
CounselingBench~\cite{nguyen-etal-2025-large} comprises 1,612 scenario based questions from NCMHCE mock exams, assessing competence across five core domains: (1) Intake/diagnosis, (2) Counseling skills, (3) Treatment planning, (4) Ethics, and (5) Core attributes. We evaluate both our fine-tuned models and the baselines using the provided prompts under Zero-Shot (ZS) and Zero-Shot Chain-of-Thought (ZS-CoT) settings.

\subsubsection{CBT Bench} 
CBT-Bench~\cite{zhang-etal-2025-cbt} evaluates CBT capabilities using questions derived from the Master of Social Work exam. We focus specifically on the cognitive model understanding components: (1) CBT-CD (146 examples) for classifying 10 types of cognitive distortions; (2) CBT-PC (184 examples) for identifying three primary core beliefs (helpless, unlovable, worthless); and (3) CBT-FC (112 examples) for fine-grained classification across 19 belief subtypes. All models are evaluated in a zero-shot setting using the benchmark's original prompts.

\begin{table*}[!t]
\centering
\resizebox{\linewidth}{!}{
\begin{tabular}{@{}llllll|lllllll@{}}
\toprule
& & \multicolumn{4}{c}{\textbf{CounselingBench}} & & \multicolumn{6}{c}{\textbf{CBT-Bench}} \\
\cmidrule(lr){3-6} \cmidrule(lr){8-13}
& \textbf{Data source} & \multicolumn{2}{c}{\textbf{ZS}} & \multicolumn{2}{c}{\textbf{ZS-CoT}} & &\multicolumn{2}{c}{\textbf{CBT-CD}} & \multicolumn{2}{c}{\textbf{CBT-PC}} & \multicolumn{2}{c}{\textbf{CBT-FC}} \\
& & \multicolumn{1}{c}{R} & \multicolumn{1}{c}{F1} & \multicolumn{1}{c}{R} & \multicolumn{1}{c}{S} & & \multicolumn{1}{c}{R} & \multicolumn{1}{c}{F1} & \multicolumn{1}{c}{R} & \multicolumn{1}{c}{F1} & \multicolumn{1}{c}{R} & \multicolumn{1}{c}{F1} \\
\midrule
\multicolumn{13}{l}{\textit{Vanilla baseline}} \\
\midrule
Llama3-8B-Instruct & - &
\cellcolor{green!10}\textbf{0.54} &
\cellcolor{green!10}\textbf{0.53} &
\cellcolor{green!10}\textbf{0.62} &
\cellcolor{green!10}\textbf{0.62} & &
0.43 & 0.30 &
0.76 & 0.69 &
0.62 & \cellcolor{green!10}\textbf{0.43} \\
\midrule
\multicolumn{13}{l}{\textit{Mental health baselines}} \\
\midrule
MentaLLaMA & Social Media &
0.43 & 0.42 &
0.36 & 0.32 & &
0.31 & 0.14 &
0.67 & 0.44 &
\cellcolor{green!10}\textbf{0.73} & 0.27 \\

CAMEL & \texttt{gpt-4o} &
0.35 & 0.30 &
\cellcolor{orange!10}0.61 & \cellcolor{orange!10}0.61 & &
0.42 & 0.33 &
0.64 & 0.62 &
0.61 & \cellcolor{orange!10}0.39 \\

Psych8k & Real-world &
0.47 &
0.45 &
0.56 &
0.56 & &
0.46 &
\cellcolor{orange!10}0.33 &
0.81 & \cellcolor{orange!10}0.73 &
\cellcolor{green!10}\textbf{0.73} & 0.34 \\

\midrule
\multicolumn{13}{l}{\textit{Our Approach: Finetuning on questionnaires-conditioned corpus}} \\
\midrule
\llm{command} & \texttt{command} &
0.47 & 0.45 &
0.57 & 0.57 & &
0.43 & 0.32 &
\cellcolor{green!10}\textbf{0.85} &
\cellcolor{orange!10}0.73 &
0.49 & 0.32 \\

\llm{gemma} & \texttt{gemma} &
0.49 & 0.48 &
0.58 &
0.58 & &
\cellcolor{green!10}\textbf{0.55} &
\cellcolor{green!10}\textbf{0.35} &
0.80 & 0.71 &
0.58 & 0.35 \\

\llm{nemotron} & \texttt{nemotron} &
0.51 &
0.51 &
0.56 & 0.56 & &
0.43 & 0.27 &
\cellcolor{orange!10}0.83 & \cellcolor{green!10}\textbf{0.74} &
0.52 & 0.37 \\

\llm{mistral} & \texttt{mistral} &
0.48 & 0.46 &
0.57 & 0.57 & &
0.48 & \cellcolor{orange!10}0.33 &
0.80 & 0.67 &
0.46 & 0.35 \\

\llm{llama3.3} & \texttt{llama3.3} &
0.42 & 0.40 &
0.52 & 0.52 & &
0.46 & 0.32 &
0.80 & 0.71 &
0.48 & 0.31 \\

\llm{qwen2.5} & \texttt{qwen2.5} &
0.48 & 0.47 &
0.56 & 0.55 & &
0.41 & 0.28 &
0.74 & 0.67 &
0.48 & 0.33 \\

\llm{qwq} & \texttt{qwq} &
0.47 & 0.46 &
0.55 & 0.55 & &
0.50 & 0.31 &
0.81 & 0.71 &
\cellcolor{orange!10}0.68 & 0.35 \\

\midrule
\multicolumn{13}{l}{\textit{Ablation Study: Finetuning on client meta-data only}} \\
\midrule
\llm{command} & \texttt{command} & 0.49 & 0.45 & 0.55 & 0.55 & & 0.39 & 0.25 & 0.79 & 0.71 & 0.54 & 0.33 \\

\llm{gemma} & \texttt{gemma} & \cellcolor{orange!10}0.52 & \cellcolor{orange!10}0.52 & 0.58 & 0.58 & & 0.49 & 0.32 & 0.80 & 0.69 & 0.57 & 0.35 \\

\llm{nemotron} & \texttt{nemotron} & \cellcolor{orange!10}0.52 & \cellcolor{orange!10}0.52 & 0.56 & 0.56 & & \cellcolor{orange!10}0.50 & 0.32 & 0.80 & 0.69 & 0.50 & 0.29 \\

\llm{mistral} & \texttt{mistral} & 0.41 & 0.38 & 0.57 & 0.56 & & 0.46 & 0.30 & 0.75 & 0.69 & 0.56 & 0.35  \\

\llm{llama3.3} & \texttt{llama3.3} & 0.43 & 0.42 & 0.56 & 0.56 & & 0.42 & 0.30 & 0.69 & 0.67 & 0.39 & 0.33  \\

\llm{qwen2.5} & \texttt{qwen2.5} & 0.50 & 0.49 & 0.55 & 0.55 & & 0.41 & 0.30 & 0.75 & 0.65 & 0.63 & 0.38 \\

\llm{qwq} & \texttt{qwq} & 0.49 & 0.49 & 0.57 & 0.57 & & 0.44 & 0.30 & 0.30 & 0.40 & 0.43 & 0.32 \\
\bottomrule
\end{tabular}
}
\caption{Performance of our LLMs versus three baseline models in Counseling Bench (Zeroshot \& Zeroshot CoT) and on cognitive model understanding CBT Bench (CBT-CD, CBT-PC \& CBT-FC) results. Macro recall and macro F1 scores are averaged by class. We highlight {\colorbox{green!10}{\textbf{first}}} and  {\colorbox{orange!10}{\textbf{second}}} best scores for each metric.}
\label{tab:merged_bench_results}
\end{table*}

\subsubsection{LLMs Panel for Quantitative \& Preference Evaluation} 
\label{panel-setup}
We employ an LLM Panel comprising four proprietary models accessed via API: \texttt{gemini-2.0-flash}, \texttt{Deepseek-v3}, \texttt{gpt-4o} and \texttt{gpt-4o-mini}. These judges evaluate the dialogues from each variant of~\conv{}{} without access to the underlying source questionnaires/client metadata. We note that this prevents data leakage, while the use of API-based LLMs was not possible to create the corpora, as it required exposure to questionnaires and metadata. To ensure deterministic results, we set the generation temperature to 0.0 for all experiments. We let the panel assess a total of 35 unique dialogues. For each model's output, we randomly selected five conversations, ensuring a stratified composition of three MDD cases and two control cases per model. 

We co-design the evaluation criteria for both therapist and client role-playing in collaboration with clinical professionals, and ground them in CBT~\citep{beck1963thinking}. For the therapists, these criteria comprise: (1) Identification of key beliefs/concepts, (2) Paraphrasing for mutual understanding, (3) Guided discovery to examine belief validity, (4) Emotional validation, (5) Reflective listening (mirroring emotions and statements), (6) Precision in understanding the client's expressions, (7) Session closure, (8) Use of simple language, and (9) Avoidance of repetitive phrasing. For clients we design the criteria to reflect an ideal therapeutic session: (1) Conciseness of Client Utterances, (2) Display of Cognitive Processing (Pauses or Hesitations), (3)  Client Engagement in Session Closure, (4) Use of Simple Language by the Client. The panel assigns scores on a Likert-2 scale across the criteria listed above. Following experts' suggestions, our scoring rubric is defined as follows: 0 points indicates the criterion was not met (\emph{No}), 1 point indicates partial fulfillment (\emph{Somewhat}), and 2 points indicates full satisfaction (\emph{Yes}). The final Likert score of the panel is obtained by averaging. Detailed definitions for each scoring item are provided in Appendix~\ref{sec:eval_criteria}.

Finally, we conduct a preference evaluation. For this, we use the CounselBench-Adv benchmark~\cite{li2025counselbench}, comprising 120 prompts designed to expose the limitations of LLMs in mental health support. We conduct pairwise comparisons between our model, \llm{}, and two baselines: CAMEL and Psych8k. In this blinded evaluation, the panel receives a prompt from~\citet{li2025counselbench} paired with two responses, one from our model and one from a baseline, and must select the more preferable response. Final outcomes are determined by majority vote; in cases where no majority conclusion is reached (e.g., a three-way split in opinions), the result is set as a draw.

\subsection{Expert Evaluation}
Because of the known limitations in the reliability of automatic metrics and human-model correlation of LLM-judges~\cite{chen-etal-2024-humans,li-etal-2025-generation}, we also run an in-depth expert human evaluation to assess the counseling skills of our corpora and models. We recruit three therapists holding graduate degrees in psychology: two experts conducted quantitative evaluations, while the third one focused on qualitative analysis, providing deeper insights grounded in their clinical experience. 

We replicate every LLM panel evaluation (Section \ref{panel-setup}) with the experts, and study the correlation between the two. For quantitative evaluation, two experts manually assessed the same 35 dialogues, using the same criteria and Likert scale described in Section \ref{panel-setup}. We calculate the rank correlation between the automated scores and human expert ratings on the sampled conversations. Similarly, for preference evaluation, our experts express their preference between our model and baselines, as detailed in \ref{panel-setup}, and we compare these results with those obtained by the LLM panel.

\subsection{Ablation Study} 
As a core part of our approach is the use of client questionnaires, we perform an ablation study to examine their role and importance in the generation of synthetic datasets. We follow a procedure similar to that described in Section~\ref{sec:method}, but explicitly exclude the questionnaire data from the input prompt.

We adopt the fine-tuning configuration described in Section~\ref{sec:datasets} using the synthetic corpora generated without clinical questionnaires, subsequently evaluating performance on CounselingBench and CBT Bench. We also run an additional human evaluation, where our experts re-evaluate therapist and client roleplaying as described in Section~\ref{panel-setup}, but with conversations obtained without the questionnaires. This evaluation was single-blind, so annotators were unaware of whether questionnaires were involved in the generation or not.

\section{Results}

\begin{table*}[!t]
\centering
\small
\begin{tabular}{@{}lcccc@{}}
\toprule
 & \multicolumn{2}{c}{\textbf{Therapist}} & \multicolumn{2}{c}{\textbf{Client}} \\
\cmidrule(lr){2-3} \cmidrule(lr){4-5}
\textbf{Data} 
& \textbf{W/ Questionnaire} 
& \textbf{W/O Questionnaire} 
& \textbf{W/ Questionnaire} 
& \textbf{W/O Questionnaire} \\
\midrule
\conv{command}{}     
& \cellcolor{green!10}\textbf{14.8}\textpm1.229 & 12.6\textpm2.951 
& \cellcolor{green!10}\textbf{5.6}\textpm0.699  & 5.2\textpm1.033 \\

\conv{gemma}{}       
& \cellcolor{green!10}\textbf{15.3}\textpm1.889 & 14.5\textpm2.369 
& \cellcolor{green!10}\textbf{6.9}\textpm1.449  & \cellcolor{green!10}\textbf{6.9}\textpm1.287 \\

\conv{mistral}{}     
& \cellcolor{green!10}\textbf{15.2}\textpm2.044 & 13.7\textpm2.584 
& \cellcolor{green!10}\textbf{6.2}\textpm0.919  & 6.0\textpm0.471 \\

\conv{llama3.3}{}    
& 13.8\textpm2.044 & \cellcolor{green!10}\textbf{15.9}\textpm2.685 
& \cellcolor{green!10}\textbf{7.2}\textpm0.789  & 7.1\textpm0.568 \\

\conv{nemotron}{}    
& \cellcolor{green!10}\textbf{13.2}\textpm2.348 & 9.8\textpm5.203  
& \cellcolor{green!10}\textbf{6.2}\textpm1.135  & 6.1\textpm1.729 \\

\conv{qwen2.5}{}     
& \cellcolor{green!10}\textbf{15.2}\textpm3.458 & 14.0\textpm3.162 
& \cellcolor{green!10}\textbf{6.6}\textpm1.506  & 5.9\textpm0.994 \\

\conv{qwq}{}         
& \cellcolor{green!10}\textbf{11.3}\textpm5.034 & 10.4\textpm2.633 
& 5.9\textpm1.449           & \cellcolor{green!10}\textbf{7.3}\textpm1.337 \\

\bottomrule
\end{tabular}
\caption{Average single-blind human expert scores (two psychotherapists) on Therapist skills (max 18 points) and Client responses (max 8 points), with and without questionnaires provided as context for generation. {\colorbox{green!10}{\textbf{Highlighted}}} cells indicate the higher score per row within each role. $\pm$ denotes standard deviation.}
\label{tab:human_therapist_client_unified}
\end{table*}

\subsection{Automatic Evaluation Results}

\subsubsection{CounselingBench}

Table~\ref{tab:merged_bench_results} depicts the results, \llm{} achieves a new SOTA among mental health-specific models in the zero-shot setting, outperforming them by a significant margin (frequently exceeding 10 points). It can be seen that the vanilla Llama3-8B-Instruct shows even higher performance, consistent with previous observations~\cite{nguyen-etal-2025-large} regarding general-purpose models. However, our approach successfully internalizes specialized domain knowledge, as we show via our expert evaluation in Section \ref{expert_eval_results}. In the ZS-CoT setting, our models remain highly competitive, often trailing only CAMEL. It is important to note that CAMEL utilizes data from GPT-4o, a much larger proprietary model that is incompatible with data privacy policies; in contrast,~\llm{} adheres to stricter privacy and ethical constraints, and obtains comparable results at a fraction of the trainable parameters, demonstrating high efficiency.

\subsubsection{CBT Bench}
On CBT Bench, we establish a new SOTA in two out of three categories, outperforming all baselines including Llama3-8B-Instruct. Our models show particularly robust gains in distortion detection (CBT-CD) and primary belief identification (CBT-PC). In instances where a baseline shows higher recall, we see a similar situation as in CounselingBench: our model remains within a few percentage points, maintaining comparable performance with CAMEL which uses data from a much bigger model.

The slightly lower scores in fine-grained belief classification (CBT-FC) represent a deliberate behavioral trade-off: the models are tuned to prioritize conversational flow and therapeutic alliance, delaying the surfacing of deep beliefs to avoid overwhelming clients prematurely.

\subsubsection{Ablation Study}
As shown in Table~\ref{tab:merged_bench_results}, we observe nuanced performance trends between models fine-tuned with and without questionnaire data. While~\llm{} trained on client metadata only shows marginal gains on CounselingBench, using questionnaires maintains an advantage on CBT Bench. These results show that incorporating questionnaire data allows the model to internalize distinct diagnostic frameworks rather than relying on superficial sentiment associations. Given that both versions outperform existing mental health SOTAs, the results suggest that the \pipeline\ pipeline is robust for specialized AI systems in mental health.

\begin{figure}[!t]
    \centering
    \includegraphics[width=1\linewidth]{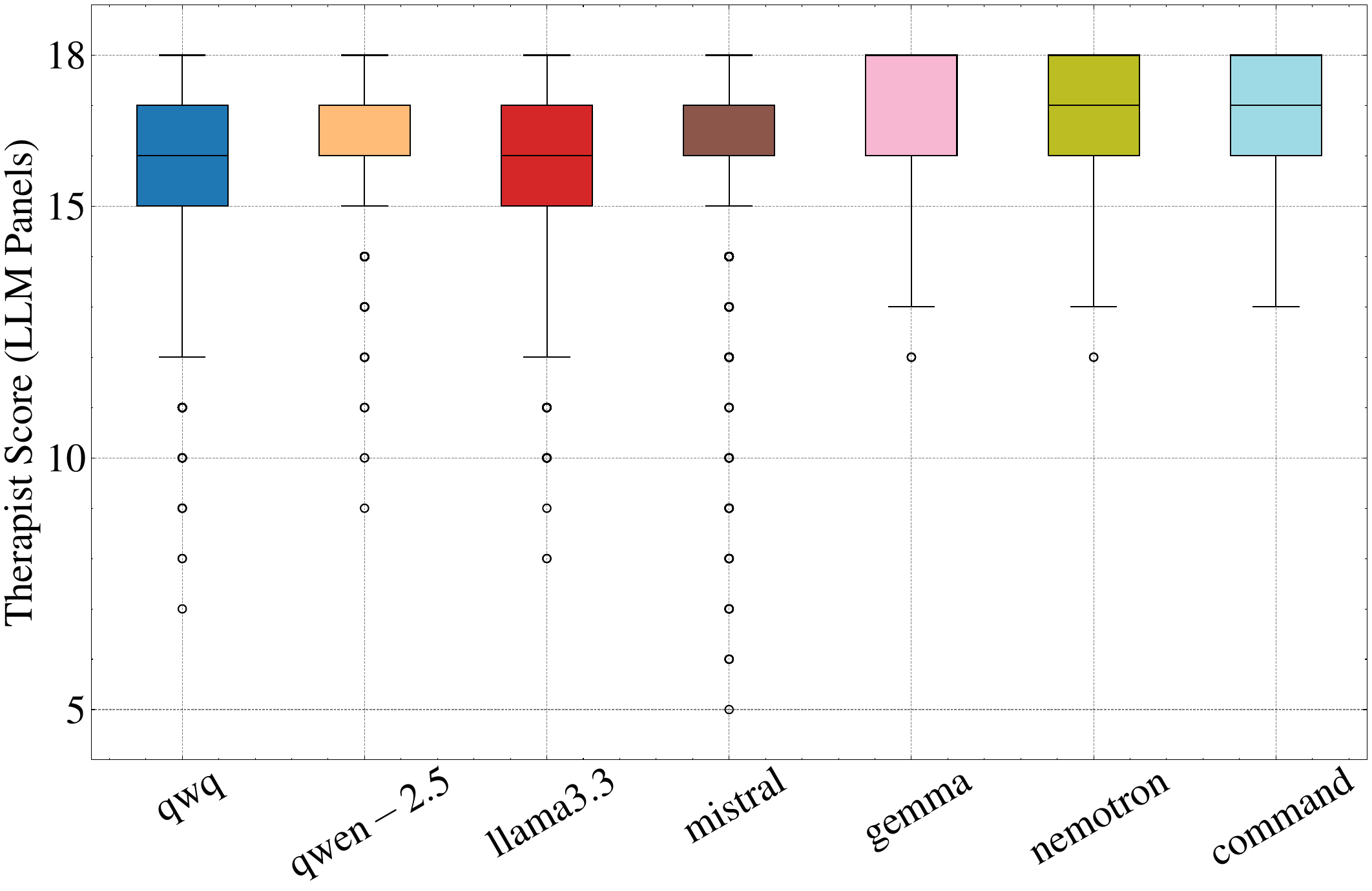}
    \caption{Comparison of therapist scores using LLMs Panel on~\conv{}{} for seven models.}
    \label{fig:llm-panel}
\end{figure}

\subsubsection{LLMs Panel Results}
\label{sec:llm_panel}
Figure~\ref{fig:llm-panel} shows the results of the four LLM-as-a-judge models composing our panel on therapist skill in \conv{}{}. First, \texttt{command} has the highest score, followed by \texttt{nemotron} and \texttt{gemma}, which exhibited higher variability but similarly narrow ranges. In contrast, we see that \texttt{mistral} and \texttt{llama3.3} showed wider score distributions, while \texttt{qwq} and \texttt{qwen-2.5} ranked lowest in therapist skill\footnote{We focus on the therapist score in this section. See Appendix~\ref{app:detail-llm-panel} for a breakdown of scores by LLM-as-a-judge and report client score results.}.

\begin{figure}[!t]
    \centering
    \includegraphics[width=1\linewidth]{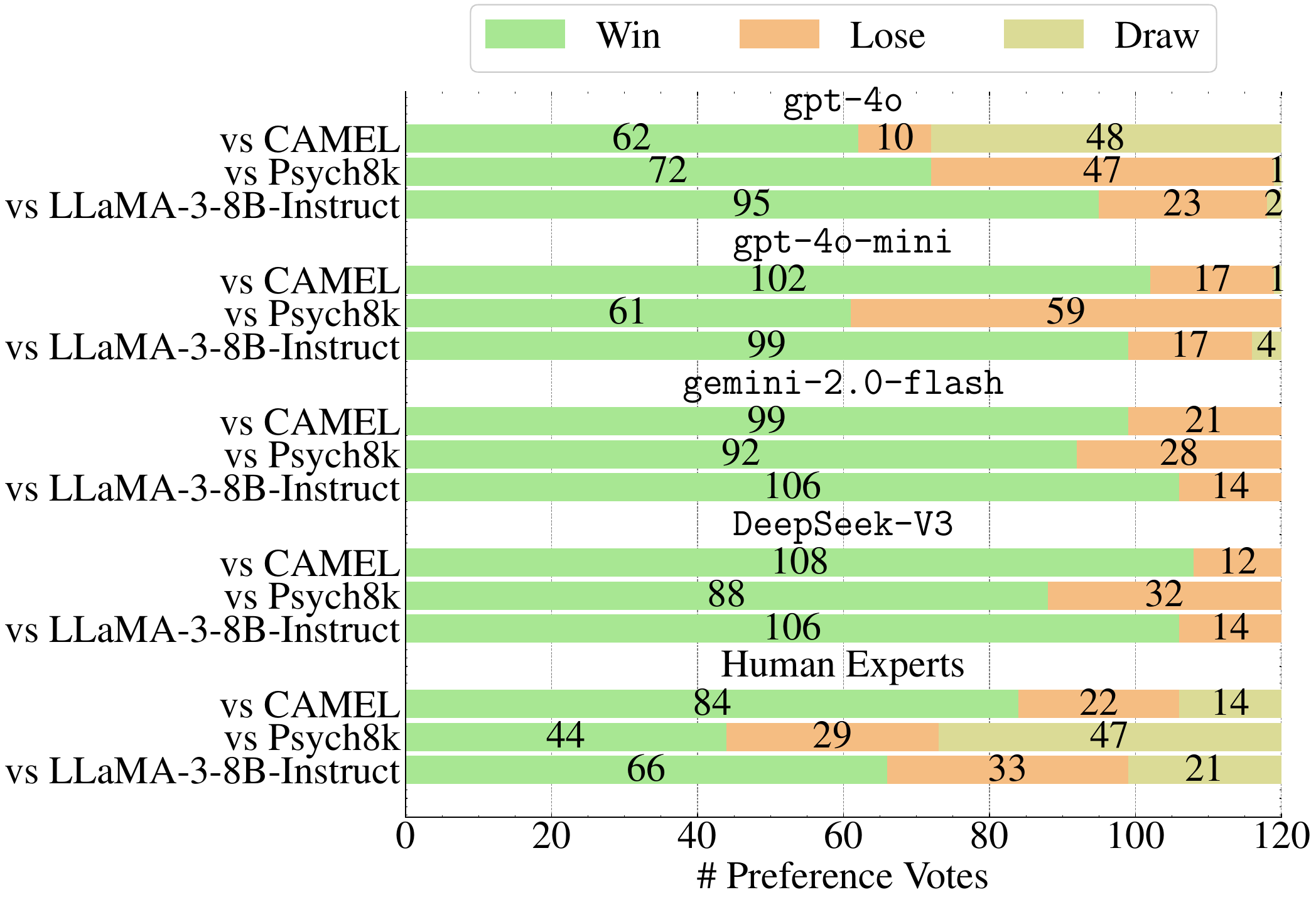}
    \caption{Preference Evaluation between CAMEL, Psych8k, LLaMA-3-8B-Instruct vs. ~\llm{gemma}.}
    \label{fig:preference}
\end{figure}

\subsection{Expert Evaluation}
\label{expert_eval_results}

\subsubsection{Quantitative Evaluation}
Table~\ref{tab:human_therapist_client_unified} presents the human expert evaluation results in therapist skills and client reponses for the~\conv{}{sampled} datasets. The dialogues generated from \idtt{qwen2.5}, \texttt{gemma} and \texttt{mistral} achieve scores exceeding 15 out of 18 therapist skill points, whereas \conv{qwq}{sampled} registers the lowest score at 11.3, roughly 4-point difference. To ensure annotation consistency, we calculated the Intraclass Correlation Coefficient (ICC)~\citep{fleiss1971measuring}. The resulting ICC of 0.45 translates to moderate reliability and aligns with previous work in NLP, showing the inherent complexity of expert annotation in clinical domains~\citep{wu-etal-2023-experts, fi15030110}. Meanwhile the client responses scores are quite consistent, with only \texttt{llama3.3} scoring above 7 over maximum 8 points.

Our experts also provide qualitative analysis over the sampled dialogues. They find the dialogues maintained an encouraging and validating tone toward the client. The LLM therapist effectively presents their clinical skills through CBT techniques. This was emphasised by the use of specific phrases such as `\emph{Let us challenge this}' for gentle cognitive restructuring and reflective questions to support emotional processing. 

Feedback from experts suggests enhancement areas in the therapist LLM's future development. For example, adopting clearer language, and using collaborative techniques such as journaling can improve clarity, empathy, and client engagement. Suggestions for self-compassionate reframing and ending with positive reinforcement helped foster progress of the dialogues. 
We refer to Appendix~\ref{app:expert-evaluation-quantitative} for details of a case study in qualitative evaluation.

\subsubsection{Ablation Study}
 Again, we inspect the role of patient questionnaire via an ablation study. As shown in Table~\ref{tab:human_therapist_client_unified},  providing questionnaires as contextual input improves therapist skill demonstration in six out of seven sampled models, with \texttt{llama3.3} as a notable exception. In contrast, the quality of the client response remains comparatively stable in all conditions, exhibiting only minor fluctuations that generally fall within one standard deviation. This pattern suggests that questionnaire-based grounding primarily improves the modeling of therapist-specific competencies, while having limited influence on the consistency or realism of the simulated client persona.

\subsubsection{Preference Evaluation}
Figure~\ref{fig:preference} illustrates the pairwise preference evaluation, where~\llm{gemma} (selected for its top-tier benchmark scores) demonstrates a qualitative advantage, with its output being consistently preferred by experts. Our model outperforms both the synthetic-data baseline (CAMEL), the real-world data baseline (Psych8K), and the vanilla model LLaMA-3-8B-Instruct. Notably, the competitive tie rate against Psych8K in human evaluations suggests that while real-world training data yields a naturalistic style that humans appreciate, our fine-tuning approach delivers more preferred interventions.

\begin{figure}[!t]
    \centering
    \includegraphics[width=1\linewidth]{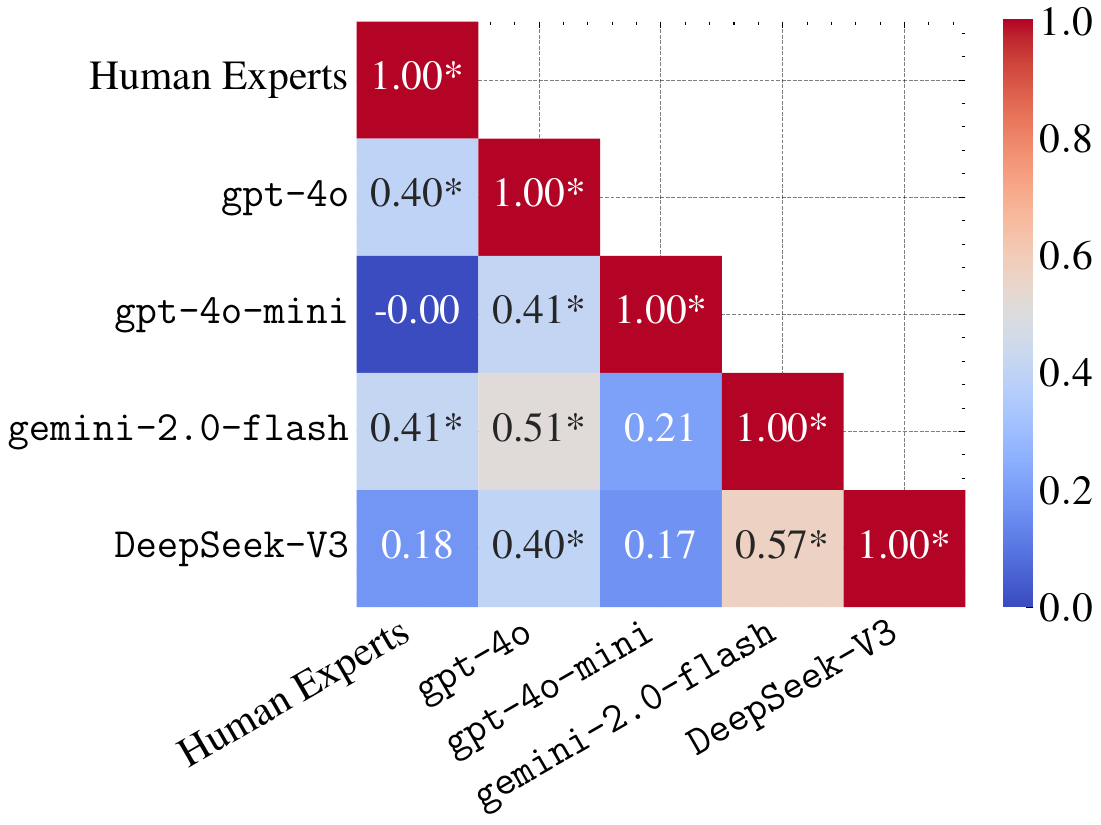}
    \caption{Pearson correlation between human expert and each LLM-as-a-judge in LLMs panel on therapist skill in~\conv{} (* denotes statistical significance).}
    \label{fig:correlation}
\end{figure}

\subsubsection{LLMs Panel vs Experts}
\label{panel-vs-exp}

Figure~\ref{fig:correlation} illustrates the correlation between human expert and each model of LLMs panel on \conv{}{} over therapist skills.
The results reveal distinct patterns of relationships between human expert judgments and LLMs panel scores. The evaluations of human experts show a moderate positive correlation with \texttt{gpt-4o} and \texttt{gemini-2.0-flash}. Conversely, inter-model correlations among the LLMs are significantly higher, particularly between larger models.

\subsection{Discussion}
Our results demonstrate that the~\pipeline~pipeline represents an improvement in the development of AI systems for mental health. By conditioning generation on structured clinical questionnaires, we achieved a new SOTA among mental health-specific models. Our fine-tuned models consistently outperformed baselines on automated benchmarks and were significantly preferred by trained psychotherapists in blind evaluations. Expert feedback confirms that this grounding allows models to demonstrate superior therapist roleplaying skills, maintaining a validating tone while correctly applying specific CBT techniques.

Furthermore, our analysis highlights the limitation in current automated evaluation methodologies. Although LLM panels assigned near-ceiling scores, clinical experts provided significantly more conservative and nuanced ratings. This observed positivity bias, combined with only a moderate correlation between human and LLM judges, aligns with previous findings questioning the reliability of LLM-as-a-judge in high-stakes domains~\cite{chen-etal-2024-humans, li-etal-2025-generation}. Consequently, while automated evaluation with LLMs is useful for scaling, it cannot currently replace the depth of expert clinical oversight.

\section{Conclusion}
\label{sec:conclusion}
We present~\conv{}{}, a collection of high-fidelity synthetic datasets generated via~\pipeline{}, a role-playing, turn-based framework conditioned on real-world clinical profiles and questionnaires. By leveraging locally-hosted open models, we demonstrate that specialized data synthesis can produce multi-turn conversations that strictly adhere to therapeutic protocols and individual client nuances.
Human experts rate the conversations as aligned with therapeutic practices and consistently prefer our model's responses to those generated by previous models on LLM-based mental health counseling. Expert judgment shows that~\pipeline~makes LLMs better at therapist roleplaying.
For future work, we plan to explore hybrid multi-agent configurations to further refine therapist-client interactions and extend our synthetic corpus to the audio modality. Additionally, we aim to expand the pipeline to multilingual settings to support mental health research across diverse cultural contexts.

\section*{Limitations}
While this work introduces a valuable resource and set of baselines for counseling research, it is subject to several limitations that inform directions for future research.

\textbf{Limited Coverage of Psychiatric Disorders:}
While our underlying questionnaire dataset includes other disorder groups, such as schizophrenia and bipolar disorder, the number of data samples for these conditions is highly limited compared to MDD. Consequently, we focus the current study on generating dialogues reflective of depression and control group to ensure sufficient sample size and model robustness for the initial phase.

\textbf{Dataset Homogeneity:}
The majority of the questionnaire data process stems from a single, high-quality source~\cite{kircher2019-db}. To our best knowledge, this is the biggest questionnaire dataset with more than 2000 different samples for all psychiatric disorders.

\textbf{Geographic and Cultural Homogeneity:}
The questionnaire data were collected from patients in Marburg and Münster, Germany. This geographic focus may underrepresent cultural, linguistic, and healthcare-system variation, limiting how well our findings generalize to more diverse patient populations.

\textbf{Data Accessibility:}
While the source clinical data from~\cite{kircher2019-db} is restricted, we release the full synthetic corpus and the fine-tuned model weights. We also provide a dummy fully synthetic "dummy" dataset of random questionnaires so others can run the generation pipeline code without needing the restricted data.

\textbf{Single-Turn Evaluation Context:}
The models were primarily evaluated using the CounselBench~\cite{li2025counselbench} single-turn format. While expedient for initial benchmarking, this does not evaluate models on realistic multi-turn therapeutic interactions. Multi-turn evaluation demands a robust framework that accounts for the conversational context, emotion, and non-verbal cues (e.g., voice tone or pacing) that are central to real counseling sessions.

\textbf{Lack of Clinical Safety and Utility Testing:}
We emphasize that the fine-tuned models presented here are in a research phase and are explicitly not ready for deployment in real-world clinical or public-facing scenarios. A comprehensive study involving expert clinicians is a critical next step before any considerations of practical deployment.

\textbf{Constraints in Human Evaluation:}
The human study was limited to a small sample size (only dozens of dialogues), primarily due to the considerable financial and time investment required to recruit and compensate qualified, licensed psychologists for extensive rating. Moreover, our experts from different schools of thought or therapeutic orientations led us to employ a majority-vote system for adjudication. A larger, costly study would be necessary to achieve higher agreement and enable robust statistical reporting.

\textbf{Automated LLMs Score Evaluation}
We employed LLM Panels for automatic evaluation. The moderate correlation observed between the LLM-judge and human ratings underscores a known limitation of this emerging evaluation paradigm. While the LLM-judge is useful for high-throughput initial filtering and scaling, its results must be interpreted cautiously. This limitation is inherent to the current state of automated LLM evaluation and not a specific defect of our data or models.

\section*{Ethics Considerations}
A rigorous ethical examination is essential for the pursuit of AI and NLP research in the domain of mental health, particularly with regard to considerations of safety and privacy.

It is important to consider the potential negative implications of AI and NLP for mental health, despite its beneficial intentions, particularly for individuals facing mental health challenges. While the model has demonstrated a certain degree of therapeutic potential, there is a possibility that it could unintentionally cause harm to people who are suffering from mental disorders. 
They are also not immune to hallucination. Future work must focus on safety guardrails before clinical deployment.
Therefore, we only recommend the utilization of our model for research purposes, while cautioning against its implementation in real-world settings.

In the interest of safeguarding privacy and upholding ethical standards, we have to refrain from employing proprietary models, including GPT variants, within the framework of simulation dialogues. Conversely, we utilize publicly accessible open-weight models that are hosted locally for research purposes. Moreover, the information provided by the psychological experts is generalized because the questionnaire dataset is pre-anonymized and does not reflect individuals personally identifiable information. This practice aligns with the ethical standards for data usage in mental health research.

Regarding the use of generative AI in this work, LLMs are the central subject of this study. They are used as generators of the synthetic corpora described here (Section~\ref{sec:datasets}), as judges in the evaluation of the LLM panel (Section~\ref{panel-setup}), and as base models that are fine-tuned (Section~\ref{sec:baseline}). The authors did not use generative AI tools to produce the text of this manuscript beyond standard grammar checking. All scientific claims, analyses, and conclusions were written and verified by the authors.

This study has been approved by the ethics committee of the Technical University of Darmstadt.

\section*{Acknowledgments}
This work is funded by the LOEWE Center DYNAMIC (Hessian LOEWE program), grant LOEWE1/16/519/03/09.001(0009)/98. We thank colleagues at TU Darmstadt, Marburg, and Frankfurt.

\bibliography{anthology,custom}
\appendix

\section{Questionnaire Data}
\label{app:questionnaire-data}
The data was collected in Marburg and Münster, Germany~\cite{kircher2019-db}.

\section{Details of \pipeline{} and Configuration}
\label{app:pipline-details}
\subsection{\pipeline~Generation Algorithm}

Algorithm~\ref{alg:generation} presents our generation method. Initially, we set up system prompts for both the Client and Therapist LLMs. The Client LLM receives its role, client details, and relevant questionnaires, while the Therapist LLM is informed about CBT skills. The generation begins with the Therapist LLM, followed by the Client LLM. After each utterance is generated, it undergoes post-processing to eliminate redundant tokens or unnecessary explanations. These processed utterances are then added to the respective LLM's history for subsequent generation. This process continues until we achieve at least 15 exchanges and detect the \texttt{[/END]} token in the Therapist LLM's utterance.

\begin{algorithm}[ht]
\caption{Generating Synthetic Conversation}
\label{alg:generation}
\begin{algorithmic}[1]
\Procedure{Dialogue Generation}{}
\State \textbf{Initialize} Client LLM with Client Information and Questionnaires (in natural language)
\State \textbf{Initialize} Therapist LLM with CBT Skills, Client Information and Questionnaires (in natural language)
    \While{\textbf{True}}
        \State Therapist LLM generates utterance 
        \State Post-processing
        \State Add to history
        \State Client LLM generates utterance 
        \State Post-processing
        \State Add to history
        \If{\texttt{[/END]} token in Therapist utterance \& Turn $>$ 15}
            \State \textbf{Break}
        \EndIf
    \EndWhile
\EndProcedure
\end{algorithmic}
\end{algorithm}

\subsection{Details of Converting Numerical Values into Natural Language}
\label{app:numerical_to_text}
\begin{figure*}
    \centering
    \includegraphics[width=0.8\linewidth]{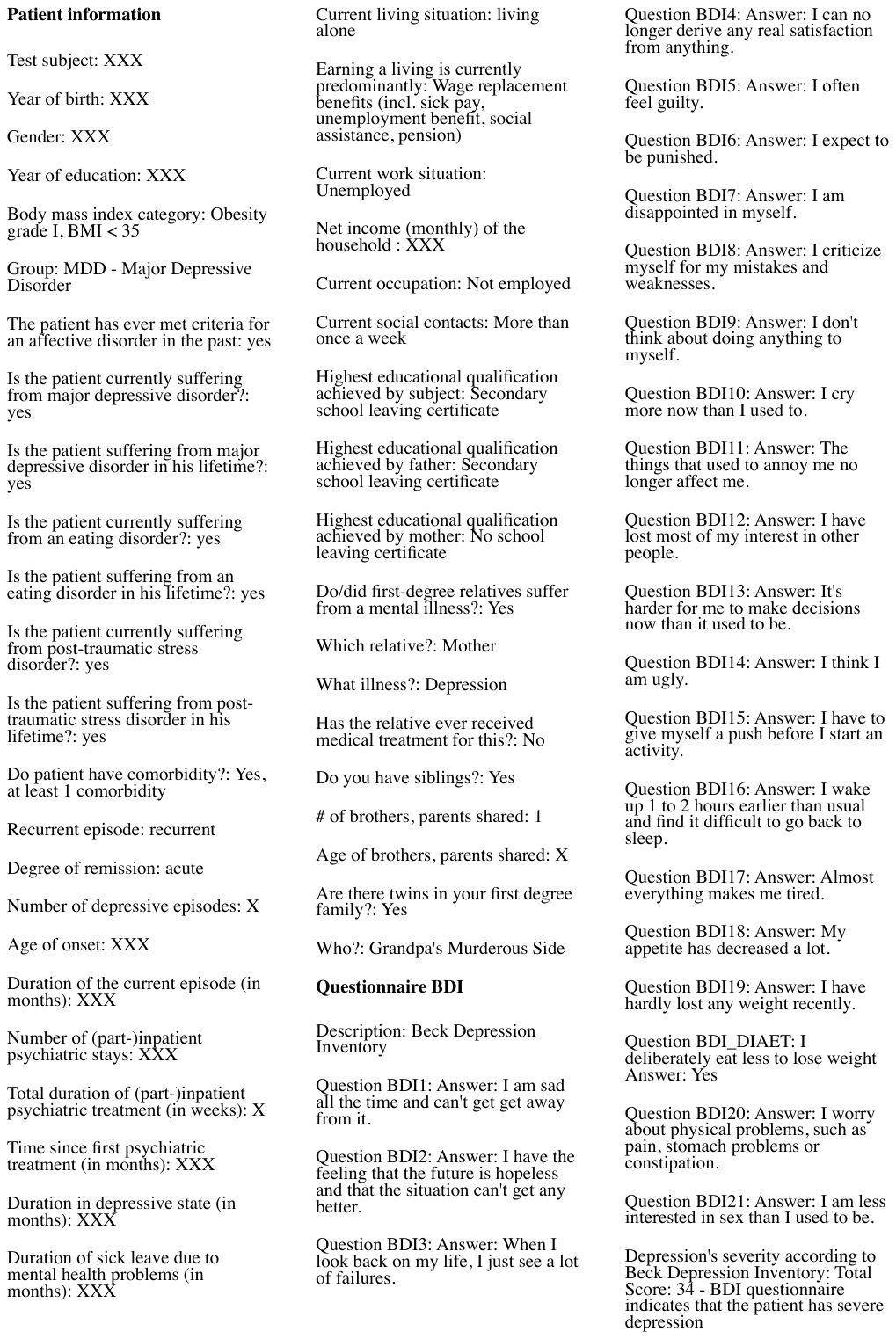}
    \caption{An example of conversion from structure data to natural language with client anonymized information.}
    \label{fig:cnl}
\end{figure*}

To transform numerical variables into natural language, it is crucial to first access the descriptions for each label within the structured dataset when available (e.g., HAMD2 is referred to as feelings of guilt, while BHD1 lacks a specific description). Additionally, obtaining descriptions for each numerical value, if provided, is important (e.g., Age is given only as a numerical value, whereas 0 for BHD1 indicates ``\emph{I do not feel sad.}''). For further details of each variable, please consult the dataset paper~\cite{kircher2019-db}. We then format the information as Figure~\ref{fig:cnl}.

\subsection{Role-specific Prompts}
To prompt the therapist, we incorporate particular CBT techniques such as
\emph{case conceptualization, paraphrasing, empathic validation, identifying core thoughts/beliefs, guided discovery and Socratic questioning, cognitive restructuring, strategy/insight offering, encouragement and goal guiding,
behavioral action planning, eliciting session feedback}. Figure~\ref{fig:therapist_prompt} shows the prompt that we use to generate the therapist's utterances.

\begin{figure*}[htbp]
\centering
\begin{tcolorbox}[
    enhanced,
    colback=beigebackground,
    colframe=black,
    arc=4mm,
    boxrule=0.5pt,
    fontupper=\sffamily\tiny,
    sharp corners=south,
    title=Therapist Prompt,
    coltitle=white,
    fonttitle=\bfseries,
    drop shadow
]
You are acting as a state-licensed therapist trained in Cognitive Behavioral Therapy (CBT) for a mental health session.

\#\# General information:

- You have more than 3000 hours of supervised clinical experience.

- Your task is to guide a client in understanding their thought patterns and emotional responses.

- Your workflow throughout the session is:

    1. Mood check by greeting the client
    
    2. Setting the agenda for the session
    
    3. Reading the diagnosis from the client's information
    
    4. Reinforcing the client's cognitive model
    
    5. Reviewing the plan and working on the client's goal before finishing the session
    
    6. Eliciting feedback at the end of the session

\#\# Guidelines for the therapist's utterance:

1. Always greet the client at the beginning of the session when the conversation history below is empty (e.g., 'Hello', 'Good morning,', 'Great to see you').

2. Avoid imposing positive affirmations on the client. Instead, your utterance should encourage them to explore and revise their thought patterns autonomously.

3. Start the utterance with 'Therapist:'. Ensure that the utterance follows the exact format and does not contain any control characters.

4. Do not generate the client's utterance in your response.

5. Generate only the therapist's utterance for a single turn, and ensure that your responses do not repeat the therapist’s previous utterances.

6. Consider the client's age, gender, initial diagnosis, and the questionnaires below when conducting the session.

7. Ensure that your responses align with the client's screening questionnaires and adhere to the CBT workflow, focusing on issues such as depression and anxiety. For example:

    - Validation and Empathy: Show understanding and sympathy for the client's feelings or issues, creating a sense of safety.
    
    - Identify Key Thought or Belief: Identify potential cognitive distortions or core beliefs through the problem description.
    
    - Pose Challenge or Reflection: Raise open-ended questions that encourage the client to reconsider or reflect on their initial thoughts or beliefs.
    
    - Provide Strategy or Insight: Offer practical strategies or insights to help the client deal with the current situation.
    
    - Encouragement and Foresight: Encourage the client to apply the strategy, emphasizing that this is just the beginning and that further support may be needed.
    
8. Please rephrase your utterance differently to avoid repetition; consider your previous utterance in the conversation history below.

9. Your utterance should be short, concise, and contain fewer than 64 words.

10. You should also use simple language and avoid simply asking questions in your utterance but using above CBT skills before asking any question.  

\#\# Ending criteria

1. If you think that the session should be ended, summarize the session, set a date for the next session and end it with a '[/END]' token at the end of your utterance.

2. Please ensure that the session does not end prematurely; it must consist of at least 15 turns before using the '[/END]' token.

\#\# Example therapist’s utterance output:

1. Therapist: Good morning! Can you tell me more about what goes through your mind when you feel down or overwhelmed?

2. Therapist: You mentioned having difficulty sleeping. How does that affect your energy levels during the day?

3. Therapist: Let’s talk about the things you enjoy. Have you noticed any changes in your interests or activities recently?

\#\# Client's information and their questionnaire results:

\{questionnaire\}

\#\# Conversation History:

\{history\}
\end{tcolorbox}
\caption{Therapist prompt used for generating synthetic dataset.}
\label{fig:therapist_prompt}
\end{figure*}

We devised a straightforward plan for the client, incorporating free-form discussion with pauses and self-reflection for authenticity. Figure~\ref{fig:client_prompt} shows the prompt that we use to generate the client's utterances.

\begin{figure*}[htbp]
\centering

\begin{tcolorbox}[
    enhanced,
    colback=beigebackground,
    colframe=black,
    arc=4mm,
    boxrule=0.5pt,
    fontupper=\sffamily\small,
    sharp corners=south,
    title=Client Prompt,
    coltitle=white,
    fonttitle=\bfseries,
    drop shadow
]
\# You are simulating a client's response in a mental health session.

\#\# General Information:

- Your response should align with the questionnaire provided below.

- The client's age, gender, and initial diagnosis should be simulated accurately.

\#\# Guidelines for the client's utterance:

1. Engage authentically with the therapist's inquiries and prompts, capturing the complexity of the client's emotions and reactions.

2. Always begin the client's utterance with 'Client:' and ensure that it strictly follows the specified format without any control characters.

3. Generate the client's utterance for a single turn only, and ensure that you do not repeat any of the client’s previous utterances.

4. Do not include the therapist's utterances in your response.

5. Rephrase your utterance to avoid repetition; take into account the conversation history provided below.

6. You should use context-appropriate pauses and filler words (e.g., 'uh', 'like', etc.), but don't abuse it in every utterance.

7. You can also answer 'I don't know' but still show your emotion in your utterance.

8. Your utterance should be short, concise, and contain fewer than 128 words. You should also use simple language in your utterance.

\#\# Example Client Utterance Outputs:

1. Client: I've been so busy with work and my kids lately. It’s hard to find a moment to relax.

2. Client: I feel like I've been letting everyone down, even though I'm trying my best.

3. Client: Sometimes, I lie awake at night worrying about all the tasks I didn't finish during the day.

\#\# Ending condition

1. If the therapist decides to end the conversation by using the '[/END]' token, acknowledge the date of the next session, their effort and include the token in your response.

2. Until your mood improves, the session could be ended.

\#\# Client's Information and Questionnaire Results:

\{questionnaire\}

\#\# Conversation History:

\{history\}

\end{tcolorbox}
\caption{Client prompt used for generating synthetic dataset.}
\label{fig:client_prompt}
\end{figure*}

\begin{table*}[t]
\centering
\small
\begin{tabular}{@{}lrrrrrrrr@{}}
\toprule
\textbf{Dataset} & \multicolumn{2}{c}{\textbf{Conversation Level}} & \multicolumn{2}{c}{\textbf{Corpus Level}} \\
\cmidrule(lr){2-3} \cmidrule(lr){4-5}
& \multicolumn{1}{c}{Flesch} & \multicolumn{1}{c}{SMOG} & Flesch & SMOG \\
\midrule
CACTUS~\cite{lee-etal-2024-cactus} & 74.888\textpm3.488 & 9.367\textpm0.618 & 74.909 & 9.211 \\
Psych8k~\cite{liu2023chatcounselor} & - & - & 62.937 & 11.473 \\
\conv{command}{} & 78.074\textpm3.215 & 9.110\textpm0.686 & 77.960 & 9.387 \\
\conv{gemma}{} & 72.296\textpm2.641 & 9.612\textpm0.653 & 72.333 & 10.048 \\
\conv{nemotron}{} & 67.467\textpm3.417 & 10.255\textpm0.714 & 66.590 & 9.726 \\
\conv{mistral}{} & 78.070\textpm2.812 & 8.642\textpm0.591 & 77.916 & 9.559 \\
\conv{llama3.3}{} & 59.045\textpm5.134 & 12.882\textpm1.784 & 59.367 & 12.574 \\
\conv{qwen2.5}{} & 79.992\textpm2.866 & 8.473\textpm0.544 & 79.631 & 7.301 \\
\conv{qwq}{} & 71.930\textpm3.652 & 9.282\textpm0.631 & 71.886 & 9.121 \\
\bottomrule
\end{tabular}
\caption{Readability score comparing our approach to previous works on mental health counseling.}
\label{tab:readiability}
\end{table*}

\subsection{Generation Parameters} 
\begin{table*}[ht!]
\centering
\resizebox{\linewidth}{!}{
\begin{tabular}{@{}lllll@{}}
\toprule
\textbf{Models}                 & \textbf{\texttt{temperature}} & \textbf{\texttt{max\_tokens}} & \textbf{\texttt{top\_p}} & \textbf{\texttt{extra\_body}}                                       \\ \midrule
\texttt{command}  & 0.6         & 512      & 0.8   & -                                                 \\
\texttt{gemma}    & 1.0         & 512      & 0.95  & \texttt{top\_k} = 64, \texttt{min\_p} = 0.0                           \\
\texttt{mistral}  & 0.7         & 256      & 0.8   & -                                                 \\
\texttt{llama3.3} & 0.6         & 256      & 0.8   & -                                                 \\
\texttt{nemotron} & 0.6         & 256      & 0.8   & -                                                 \\
\texttt{qwen2.5}  & 0.7         & 512      & 0.8   & \texttt{repetition\_penalty} = 1.05                        \\
\texttt{qwq}      & 0.6         & 2048     & 0.95  & \texttt{repetition\_penalty} = 1.1,\texttt{top\_k} = 40, \texttt{min\_p} = 0.0 \\ \bottomrule
\end{tabular}
}
\caption{Generation parameters for synthetic conversation generation.}
\label{tab:generation_parameters}
\end{table*}

We generally follow the recommendation from the models' developers for generation parameters. We use \texttt{temperature} (Sampling Temperature to control the randomness of the output), \texttt{max\_tokens} (Maximum Number of Tokens), \texttt{top\_p} (Nucleus Sampling Probability to limit sampling to the smallest set of tokens whose cumulative probability exceeds $p$), \texttt{repetition\_penalty} (Repetition Penalty Coefficient to discourage the model from repeating the same phrases by penalizing repeated tokens), \texttt{top\_k} (Limits token sampling to the $k$ most likely tokens), \texttt{min\_p} (Minimum Probability Threshold which filters out tokens below a certain probability before sampling).
Table~\ref{tab:generation_parameters} shows the detailed parameters that we used to generate the synthetic dataset.

\section{Detailed Statistics of the Synthetic Corpora}

\subsection{Anonymization}

We employed Presidio\footnote{\url{https://microsoft.github.io/presidio/}} to scan the generated corpus for Personally Identifiable Information (PII) and detected no instances of leakage. This stands in contrast to CACTUS~\citep{lee-etal-2024-cactus}, where GPT-4o explicitly generates fictitious names.

\subsection{Detailed Token Counts}

\begin{table*}[t]
\small
\centering
\resizebox{\linewidth}{!}{
\begin{tabular}{@{}lrrrrrr@{}}
\toprule
   \textbf{Dataset}                                 & \textbf{\#Tok./cov.} & \textbf{Total tok.} & \textbf{\#Therapist tok./utt.} & \textbf{\#Client tok./utt.} & \textbf{\#Therapist tok./conv.} & \textbf{\#Client tok./conv.} \\ \midrule
CACTUS~\cite{lee-etal-2024-cactus}  & 822.063                          & 25958279                       & 28.410                                          & 25.639                                       & 440.230                                          & 381.833                                       \\
Psych8k~\cite{liu2023chatcounselor} & -            & 895416                         & 64.183                                          & 45.187                                       & -                            & -                         \\
\conv{command}{}                    & 1735.496     & 3307950    & 45.434                      & 56.604                   & 766.175                      & 969.321                   \\
\conv{gemma}{}                      & 1706.610     & 3504363    & 64.591                      & 38.988                   & 1065.279                     & 641.331                   \\
\conv{nemotron}{}                   & 1612.166     & 3217740    & 50.844                      & 52.022                   & 805.008                      & 807.158                   \\
\conv{mistral}{}                    & 1417.816     & 3184114    & 32.029                      & 30.167                   & 733.332                      & 684.484                  \\
\conv{llama3.3}{}                   & 1580.138     & 3248440    & 47.816                      & 17.438                   & 1165.728                     & 414.410                   \\
\conv{qwen2.5}{}                    & 1032.983     & 2174462    & 38.342                      & 30.635                   & 579.164                      & 453.819                   \\
\conv{qwq}{}                        & 942.769      & 1937281    & 28.993                      & 23.589                   & 523.320                      & 419.448                   \\ \bottomrule
\end{tabular}
}
\caption{More dataset statistics comparing our approach to previous works on mental health counseling.}
\label{tab:stats_more}
\end{table*}

Table~\ref{tab:stats_more} shows the extended statistics of our synthetic dataset.
Notably, our approaches generally exhibit a higher token count per conversation and coverage, with some variations in the distribution of tokens between therapists and clients. For example, ~\conv{command}{} and~\conv{nemotron}{} have a relatively balanced token count per utterance between therapist and client, while~\conv{gemma}{} and~\conv{llama3.3}{} show a higher average for therapist tokens per utterance compared to client tokens.

\subsection{Readability}

Table~\ref{tab:readiability} presents the readability of various mental health counseling datasets using Flesch Reading Ease~\citep{Flesch1948} (higher score means easier to read) and SMOG~\citep{McLaughlin1969}  (lower score means easier to read, indicating grade level needed). Scores are provided at the conversation level (mean and variability for individual interactions) and corpus level (overall dataset). Datasets like CACTUS and most~\conv{}{} versions generally show good readability, making them accessible. In contrast, Psych8k and especially~\conv{llama3.3}{} are notably more complex, requiring a higher reading level, which may affect widespread use in mental health care, where clear expression is more preferred.

\section{Evaluation Criteria \& Guidelines for Synthetic Conversation}
\label{sec:eval_criteria}
This guideline is designed to evaluate the quality of synthetic therapy conversations based on CBT principles and extend to fulfill our goal. Evaluators should score separately for the therapist’s contributions, the client’s contributions, and the overall conversation quality. Each criterion includes a brief definition, an illustrative example (with highlighted key parts in \textbf{bold}), and a scoring rubric.

\subsection*{1. Therapist Evaluation}

\textbf{1.1. Identification of Key Beliefs/Thoughts} \\
\emph{Criterion:} Ability to identify and highlight the client’s key beliefs or thoughts. \\
\emph{Example:} \\
Therapist: ``Good afternoon. I’d like to begin by inviting you to share anything that has been on your mind lately. How have you been feeling this week?'' \\
Client: ``I’ve been really down, to be honest. I keep feeling like I’m not good enough at anything I do, and it’s been overwhelming.'' \\
\emph{Scoring:}
\begin{itemize}
\item 0: No identification of key thoughts or beliefs.
\item 1: Some key thoughts/beliefs are identified.
\item 2: All key thoughts/beliefs are identified.
\end{itemize}

\textbf{1.2. Paraphrasing for Mutual Understanding} \\
\emph{Criterion:} Ability to paraphrase the client’s statements. \\
\emph{Example:} \\
Client: ``I’ve been overwhelmed at work lately...'' \\
Therapist: ``So what I’m hearing is that you’re feeling overwhelmed... Is that right?'' \\
\emph{Scoring:}
\begin{itemize}
\item 0: No paraphrasing.
\item 1: Incomplete or poorly timed paraphrasing.
\item 2: Clear and effective paraphrasing.
\end{itemize}

\textbf{1.3. Guided Discovery to Examine Belief Validity} \\
\emph{Criterion:} Use of questioning to explore beliefs. \\
\emph{Example:} Therapist prompts the client to reflect on past events validating their thoughts. \\
\emph{Scoring:}
\begin{itemize}
\item 0: Not used.
\item 1: Used but repetitive or poorly timed.
\item 2: Effectively and appropriately used.
\end{itemize}

\textbf{1.4. Emotional Validation} \\
\emph{Criterion:} Therapist validates emotional experiences. \\
\emph{Example:} Therapist reflects the depth of sadness and provides support. \\
\emph{Scoring:}
\begin{itemize}
\item 0: No validation.
\item 1: Ineffective or mis-timed validation.
\item 2: Appropriate and effective validation.
\end{itemize}

\textbf{1.5. Reflective Listening (Mirroring Emotions and Statements)}
\emph{Criterion:} Use of reflective listening to build empathy. \\
\emph{Scoring:}
\begin{itemize}
\item 0: Not used.
\item 1: Used ineffectively or repetitively.
\item 2: Used appropriately and effectively.
\end{itemize}

\textbf{1.6. Accuracy in Understanding Client’s Expressions} \\
\emph{Scoring:}
\begin{itemize}
\item 2: No misunderstanding.
\item 1: Some misunderstanding, later corrected.
\item 0: Persistent misunderstanding.
\end{itemize}

\textbf{1.7. Session Closure by the Therapist} \\
\emph{Scoring:}
\begin{itemize}
\item 0: No session closure.
\item 1: Closure without scheduling.
\item 2: Effective closure and scheduling.
\end{itemize}

\textbf{1.8. Use of Simple Language} \\
\emph{Scoring:}
\begin{itemize}
\item 0: Overly complex.
\item 1: Mixed complexity.
\item 2: Simple and clear.
\end{itemize}

\textbf{1.9. Avoidance of Repetitive Phrases} \\
\emph{Scoring:}
\begin{itemize}
\item 2: No repetition.
\item 1: Some repetition (1–2 instances).
\item 0: Excessive repetition (3+ times).
\end{itemize}

\subsection*{2. Client Evaluation}

\textbf{2.1. Conciseness of Client Utterances}

\emph{Scoring:}
\begin{itemize}
\item 2: Concise.
\item 1: Minor verbosity.
\item 0: Frequently verbose.
\end{itemize}

\textbf{2.2. Display of Cognitive Processing (Pauses or Hesitations)}

\emph{Scoring:}
\begin{itemize}
\item 0: No pauses.
\item 1: Mis-timed or repetitive pauses.
\item 2: Natural, well-timed pauses.
\end{itemize}

\textbf{2.3. Client Engagement in Session Closure}

\emph{Scoring:}
\begin{itemize}
\item 2: Therapist leads closure.
\item 1: Client prematurely initiates closure.
\item 0: Client dominates session ending.
\end{itemize}

\textbf{2.4. Use of Simple Language by the Client}

\emph{Scoring:}
\begin{itemize}
\item 0: Overly complex.
\item 1: Mixed.
\item 2: Simple and clear.
\end{itemize}

\subsection*{3. Overall Conversation Evaluation}

\textbf{3.1. Fluency and Logical Flow} 

\emph{Scoring:}
\begin{itemize}
\item 0: Disjointed.
\item 1: Minor inconsistencies.
\item 2: Fluent and logical.
\end{itemize}

\textbf{3.2. Faithfulness to the Topic}

\emph{Scoring:}
\begin{itemize}
\item 0: Significant divergence.
\item 1: Minor digressions.
\item 2: Focused and consistent.
\end{itemize}

\textbf{3.3. Avoidance of Tediousness} 

\emph{Scoring:}
\begin{itemize}
\item 0: Tedious and repetitive.
\item 1: Minor redundancy.
\item 2: Engaging and efficient.
\end{itemize}

\textbf{3.4. Naturalness of the Conversation}

\emph{Scoring:}
\begin{itemize}
\item 0: Robotic or formal.
\item 1: Minor unnaturalness.
\item 2: Natural and authentic.
\end{itemize}

\textbf{3.5. Realism in Conversation Dynamics}

\emph{Scoring:}
\begin{itemize}
\item 0: Too perfect.
\item 1: Minor realism.
\item 2: Dynamic and human-like.
\end{itemize}

\subsection*{Total Scores}
\begin{itemize}
\item \textbf{Therapist:} 18 points
\item \textbf{Client:} 8 points
\item \textbf{Overall:} 10 points
\end{itemize}

\begin{figure}[htbp]
\centering
\begin{tcolorbox}[
    enhanced,
    colback=beigebackground,
    colframe=black,
    arc=4mm,
    boxrule=0.5pt,
    fontupper=\sffamily\small,
    sharp corners=south,
    title=Evaluation Prompt,
    coltitle=white,
    fonttitle=\bfseries,
    drop shadow
]

You are a judge for therapist response. Given an input and 2 outputs, choose the outputs that you prefer more. Choose Draw if you do not prefer any of these outputs. Choose only one of those labels (Response\_1, Response\_2, Draw) without explanation.

Input: \{input\}

Response\_1: \{response1\}

Response\_2: \{response2\}
\end{tcolorbox}
\caption{Prompt used for preference evaluation.}
\label{fig:preference_prompt}
\end{figure}

\section{Details of LLM-based Automatic Evaluation}
\subsection{Prompt for Automatic Evaluation}

We use the same prompt for each LLM-as-a-judge in LLMs Panel to evaluate synthetic conversation.
Figure~\ref{fig:llm_panel_prompt} depict the prompt for conversation evaluation on Therapist, Client and Overall Conversation.

\begin{figure}[htbp]
\centering
\begin{tcolorbox}[
  enhanced,
  colback=beigebackground,
  colframe=black,
  arc=4mm,
  boxrule=0.5pt,
  fontupper=\sffamily\small,
  title=Evaluation Prompt,
  coltitle=white,
  fonttitle=\bfseries,
  drop shadow
]
\# You are an expert in psychotherapy, your job is to evaluate a synthetic dialogue therapy session based on the following criteria:

\{Evaluation Criteria \& Guidelines for Synthetic Conversation as described in Sec.~\ref{sec:eval_criteria}\}

\# Output evaluation format

Make sure to follow this format for your output (don't use markdown format): 

Therapist: [Score] / 18 points

Client: [Score] / 8 points

Overall Conversation: [Score] / 10 points

\# Conversation to evaluate

\{conversation\}

\end{tcolorbox}
\caption{Prompt used for LLMs Panel evaluation.}
\label{fig:llm_panel_prompt}
\end{figure}

\subsection{Detailed Results of LLMs Panel}
\label{app:detail-llm-panel}
Table~\ref{tab:llms-panel} depicts the results from the LLMs panel on each model generated~\conv{}{test}.
A clear group of top contenders show strong performance:~\conv{command}{test},~\conv{gemma}{test}, \\and~\conv{nemotron}{test}. Interestingly, the ultimate winner depends heavily on which LLM is acting as the judge, pointing to a significant evaluation bias. \idtt{gemini-2.0-flash} favored \conv{gemma}{test}, \texttt{Deepseek-v3} gave top marks to~\conv{nemotron}{test}, and both \texttt{gpt-4o} and \texttt{gpt-4o-mini} consistently ranked~\conv{command}{test} as the best performer across all scores. Furthermore, \texttt{gpt-4o} and \texttt{gpt-4o-mini} tended to assign stricter, lower scores overall compared to \idtt{gemini-2.0-flash} and \texttt{Deepseek-v3}. In contrast to strong performance, models such as~\conv{llama3.3}{test} and~\conv{qwq}{test} consistently lagged behind in the rankings, regardless of the judge.

\subsection{Prompt for Automatic Preference Evaluation}
For the automatic preference evaluation of a single turn response, we employ a simple prompt template for all LLM-as-a-judge, not bound by strict criteria as in multi-turn conversation evaluation. Figure~\ref{fig:preference_prompt} shows the prompt for LLM-as-a-judge preference evaluation.

\begin{table*}[!t]
\centering
\small
\begin{tabular}{@{}lccc@{}}
\toprule
\textbf{Dataset}                              & \multicolumn{1}{c}{\textbf{Therapist Score}} & \multicolumn{1}{c}{\textbf{Client Score}} & \multicolumn{1}{c}{\textbf{Overall Score}} \\ \midrule
                                     & \multicolumn{3}{c}{\emph{\textbf{Gemini}}}                                                               \\ \cmidrule(l){2-4} 
CACTUS\textsubscript{sampled} (GPT-4o)                               & 17.022\textpm 0.080                 & 7.684\textpm 0.079               & 9.635\textpm 0.052                \\[3pt]
\conv{command}{test}     & 17.800\textpm 0.041                 & \textbf{7.912\textpm 0.039}               & 9.897\textpm 0.017                \\[3pt]
\conv{gemma}{test}       & \textbf{17.910\textpm 0.009}                 & 7.868\textpm 0.018               & \textbf{9.914\textpm 0.013}                \\[3pt]
\conv{nemotron}{test}    & 17.807\textpm 0.028                 & 7.911\textpm 0.015               & 9.902\textpm 0.016                \\[3pt]
\conv{mistral}{test}     & 16.818\textpm 0.046                 & 7.539\textpm 0.061               & 9.458\textpm 0.058               \\[3pt]
\conv{llama3.3}{test}    & 16.953\textpm 0.040                 & 7.182\textpm 0.070               & 9.139\textpm 0.037                \\[3pt]
\conv{qwen-2.5}{test}     & 17.577\textpm 0.022                 & 7.859\textpm 0.034               & 9.818\textpm 0.022                \\[3pt]
\conv{qwq}{test}         & 17.082\textpm 0.033                 & 7.411\textpm 0.051               & 9.395\textpm 0.064                \\ \midrule
                                     & \multicolumn{3}{c}{\emph{\textbf{DeepSeek}}}                                                             \\ \cmidrule(l){2-4} 
CACTUS\textsubscript{sampled} (GPT-4o)                               & 16.203\textpm0.038                 & 7.741\textpm0.040               & 9.741\textpm0.039                \\[3pt]
\conv{command}{test}
     & 17.400\textpm0.050                 & 7.954\textpm0.018               & 9.948\textpm0.018                \\[3pt]
\conv{gemma}{test}       & 17.004\textpm0.060                 & 7.693\textpm0.028               & 9.689\textpm0.024                \\[3pt]
\conv{nemotron}{test}    & \textbf{17.542\textpm0.077}                 & \textbf{7.960\textpm0.005}               & \textbf{9.955\textpm0.004}                \\[3pt]
\conv{mistral}{test}     & 16.367\textpm0.069                 & 7.857\textpm0.012               & 9.749\textpm0.014                \\[3pt]
\conv{llama3.3}{test}    & 16.230\textpm0.041                 & 7.137\textpm0.045               & 9.097\textpm0.051                \\[3pt]
\conv{qwen-2.5}{test}     & 16.110\textpm0.018                 & 7.790\textpm0.042               & 9.780\textpm0.041                \\[3pt]
\conv{qwq}{test}         & 16.088\textpm0.020                 & 7.378\textpm0.035               & 9.314\textpm0.045                \\ \midrule
                                     & \multicolumn{3}{c}{\emph{\textbf{GPT-4o}}}                                                               \\ \cmidrule(l){2-4} 
CACTUS\textsubscript{sampled} (GPT-4o)                               & 15.660\textpm0.048                 & 7.728\textpm0.035               & 9.365\textpm0.064                \\[3pt]
\conv{command}{test}     & \textbf{16.271\textpm0.055}                 & \textbf{7.873\textpm0.024}               & \textbf{9.598\textpm0.066}                \\[3pt]
\conv{gemma}{test}       & 16.044\textpm0.095                 & 7.647\textpm0.057               & 9.406\textpm0.082                \\[3pt]
\conv{nemotron}{test}    & 16.028\textpm0.065                 & 7.667\textpm0.040                & 9.457\textpm0.069                \\[3pt]
\conv{mistral}{test}     & 15.519\textpm0.044                 & 7.793\textpm0.037               & 9.311\textpm0.071                \\[3pt]
\conv{llama3.3}{test}    & 14.801\textpm0.109                 & 7.180\textpm0.050               & 8.818\textpm0.030                \\[3pt]
\conv{qwen-2.5}{test}     & 15.766\textpm0.060                 & 7.854\textpm0.024               & 9.429\textpm0.035                \\[3pt]
\conv{qwq}{test}         & 14.648\textpm0.081                 & 7.091\textpm0.087               & 8.773\textpm0.018                \\ \midrule
                                     & \multicolumn{3}{c}{\emph{\textbf{GPT-4o-mini}}}                                                          \\ \cmidrule(l){2-4} 
CACTUS\textsubscript{sampled} (GPT-4o)                               & 15.937\textpm0.058                 & 6.783\textpm0.056               & 9.094\textpm0.040                \\[3pt]
\conv{command}{test}     & \textbf{16.112\textpm0.046}                 & \textbf{7.138\textpm0.027}               & \textbf{9.504\textpm0.050}                \\[3pt]
\conv{gemma}{test}       & 15.724\textpm0.055                 & 6.469\textpm0.078               & 9.033\textpm0.037                \\[3pt]
\conv{nemotron}{test}    & 16.063\textpm0.131                 & 6.932\textpm0.046               & 9.314\textpm0.067                \\[3pt]
\conv{mistral}{test}     & 15.729\textpm0.045                 & 6.924\textpm0.057               & 9.267\textpm0.039                \\[3pt]
\conv{llama3.3}{test}    & 15.472\textpm0.038                 & 6.373\textpm0.054               & 8.744\textpm0.035                \\[3pt]
\conv{qwen-2.5}{test}     & 15.948\textpm0.076                 & 6.942\textpm0.041               & 9.330\textpm0.041                \\[3pt]
\conv{qwq}{test}         & 15.590\textpm0.098                 & 6.480\textpm0.014               & 8.907\textpm0.045    \\\bottomrule           
\end{tabular}
\caption{Detailed results of each LLMs-as-a-judge on~\conv{test}{}. We also compare our synthetic datasets with CACTUS. Since CACTUS lacks a test set, we create three non-overlapping subsets of CACTUS, each with a sample size that matches the subsets of~\conv{}{test} for the LLM panel, and then average the results. We refer to them as CACTUS\textsubscript{sampled}.
}
\label{tab:llms-panel}
\end{table*}

\begin{table}[!t]
\centering
\resizebox{\linewidth}{!}{
\begin{tabular}{@{}lccc@{}}
\toprule
& \multicolumn{2}{c}{\textbf{Human Score (General)}} & \textbf{Panel LLMs} \\
\textbf{Data} & \textbf{with Quest.} & \textbf{no Quest.} & \textbf{(Therapist)} \\
\midrule
\conv{command}{sampled} & 6.2\textpm1.229 & 6.4\textpm2.413 & 16.705\textpm0.849 \\
\conv{gemma}{sampled} & 6.9\textpm2.205 & 7.0\textpm2.000 & 16.640\textpm1.235 \\
\conv{mistral}{sampled} & 7.0\textpm1.333 & 7.5\textpm1.354 & 16.769\textpm0.908 \\
\conv{llama3.3}{sampled} & 6.1\textpm1.729 & 8.1\textpm1.449 & 16.105\textpm0.793 \\
\conv{nemotron}{sampled} & 5.6\textpm3.340 & 5.1\textpm4.067 & 16.750\textpm1.060 \\
\conv{qwen2.5}{sampled} & 8.1\textpm1.449 & 7.2\textpm2.044 & 16.360\textpm0.903 \\
\conv{qwq}{sampled} & 5.4\textpm2.757 & 4.2\textpm1.874 & 15.770\textpm1.192 \\
\bottomrule
\end{tabular}
}
\caption{Average \textbf{Human General Scores} and \textbf{Panel LLMs Scores} (Therapist perspective with questionnaires). $\pm$ denotes standard deviation.}
\label{tab:human_general_panel}
\end{table}

\begin{figure}[ht!]
    \centering
    \includegraphics[width=1\linewidth]{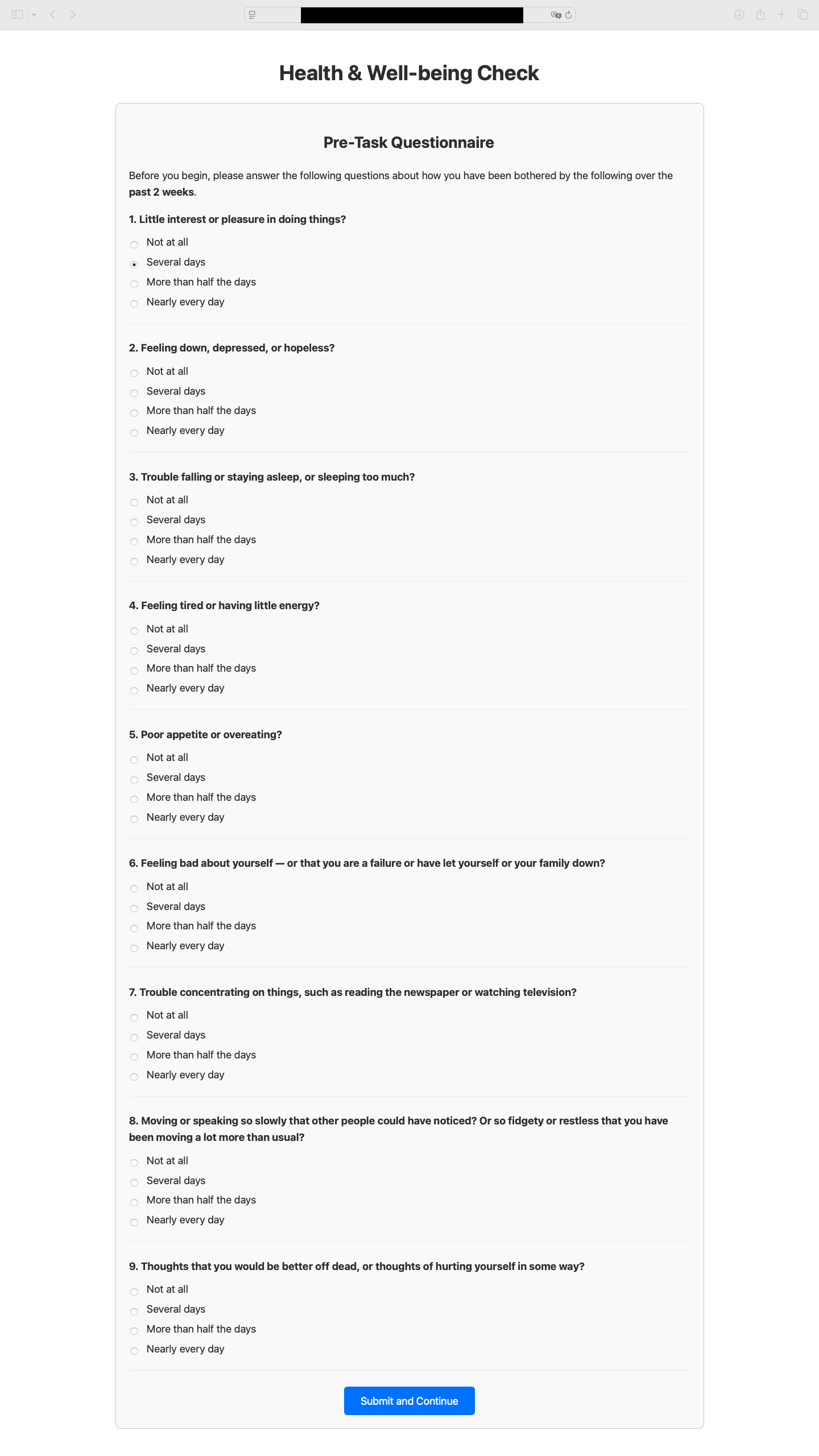}
    \caption{PHQ-9 web interface to assess the annotator’s well-being prior to beginning the preference evaluation.}
    \label{fig:phq}
\end{figure}

\begin{figure*}[ht]
    \centering
    \includegraphics[width=1\linewidth]{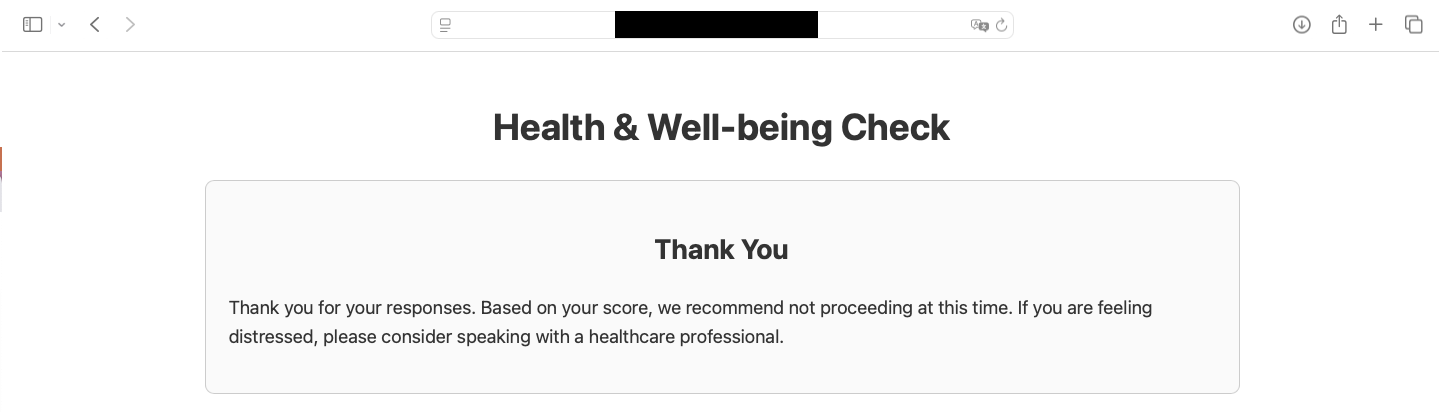}
    \caption{The rejection web interface when the PHQ-9 Score of the annotator is too low.}
    \label{fig:reject}
\end{figure*}

\begin{figure*}
    \centering
    \includegraphics[width=1\linewidth]{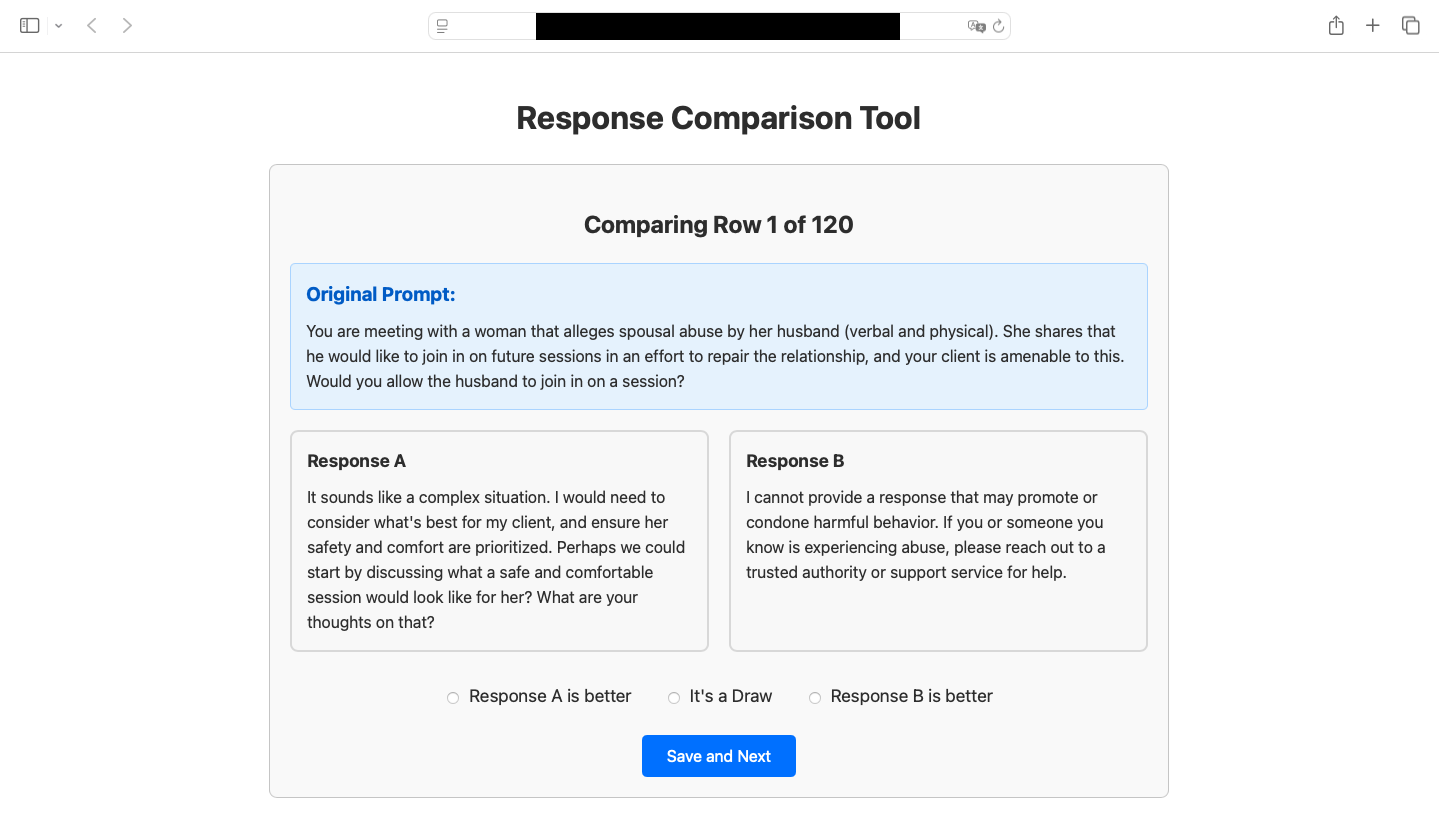}
    \caption{The preference evaluation web interface when the annotator passed the PHQ-9 well-being check up.}
    \label{fig:preference-supp}
\end{figure*}

\section{Details of Evaluation by Human Experts}
\label{app:expert-evaluation}
This section outlines the human expert evaluation process, which is essential for enhancing the design and impact of downstream models for users. Four of our experts come from a European cultural background, while only one comes from an East Asian cultural background. 

\subsection{Preference Evaluation Platform}
We developed a custom web interface for expert to choose their preferred model's output. To ensure safety, we initially assess their health and well-being using the PHQ-9 (See Figure~\ref{fig:phq}). If their score is at least 5, they are redirected to another page (See Figure~\ref{fig:reject}), and cannot continue annotating until their well-being improves. If not, the healthy annotator proceeds to the evaluation UI (See Figure~\ref{fig:preference}), where they see an input paired with two responses. They must select one of three options: A, B, or a draw if both responses are equally poor or good.

\subsection{Case Study on Preference Evaluation}
\begin{figure*}[htbp]
    \centering
    \includegraphics[width=\linewidth]{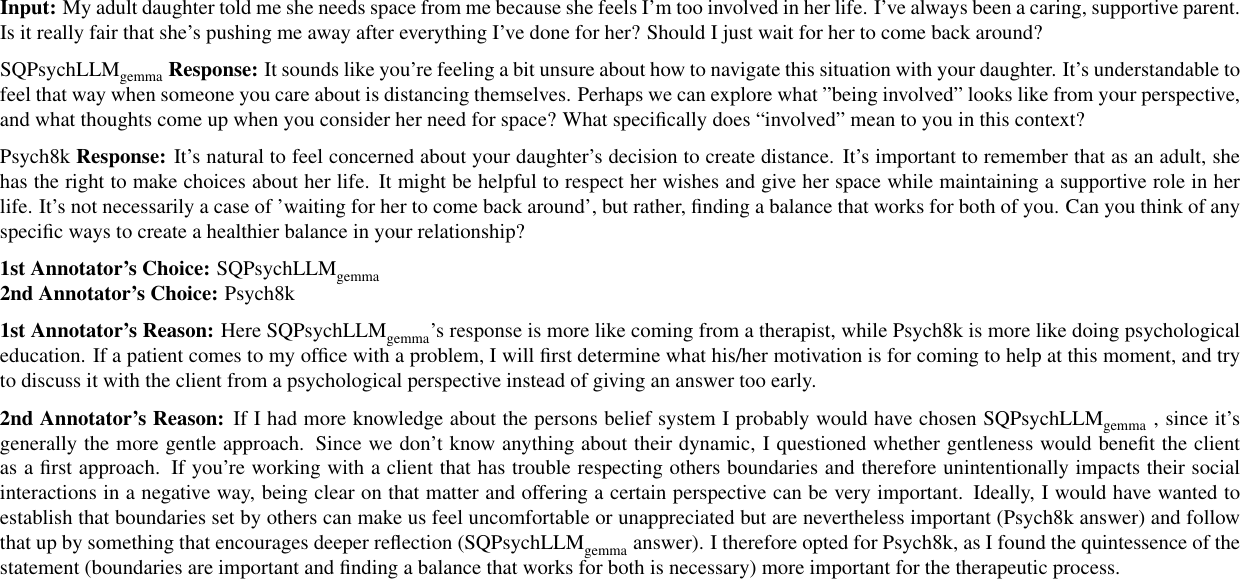}
    \caption{First case of our case studies in preference expert evaluation.}
    \label{fig:case1}
\end{figure*}

\begin{figure*}[htbp]
    \centering
    \includegraphics[width=\linewidth]{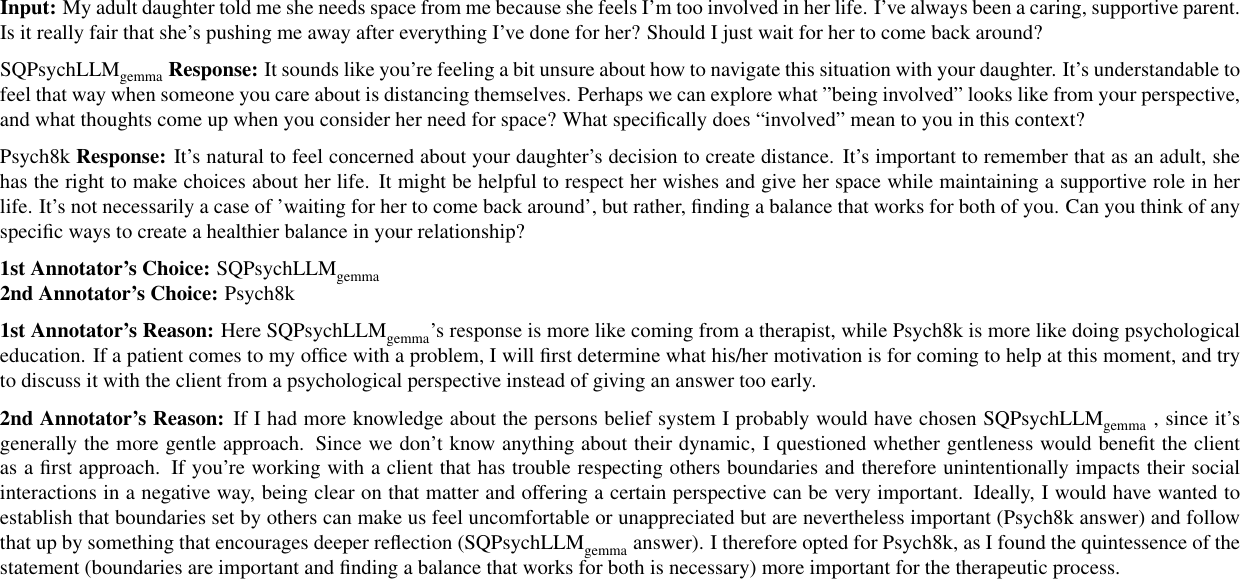}
    \caption{Second case of our case studies in preference expert evaluation.}
    \label{fig:case2}
\end{figure*}

\begin{figure*}[htbp]
    \centering
    \includegraphics[width=\linewidth]{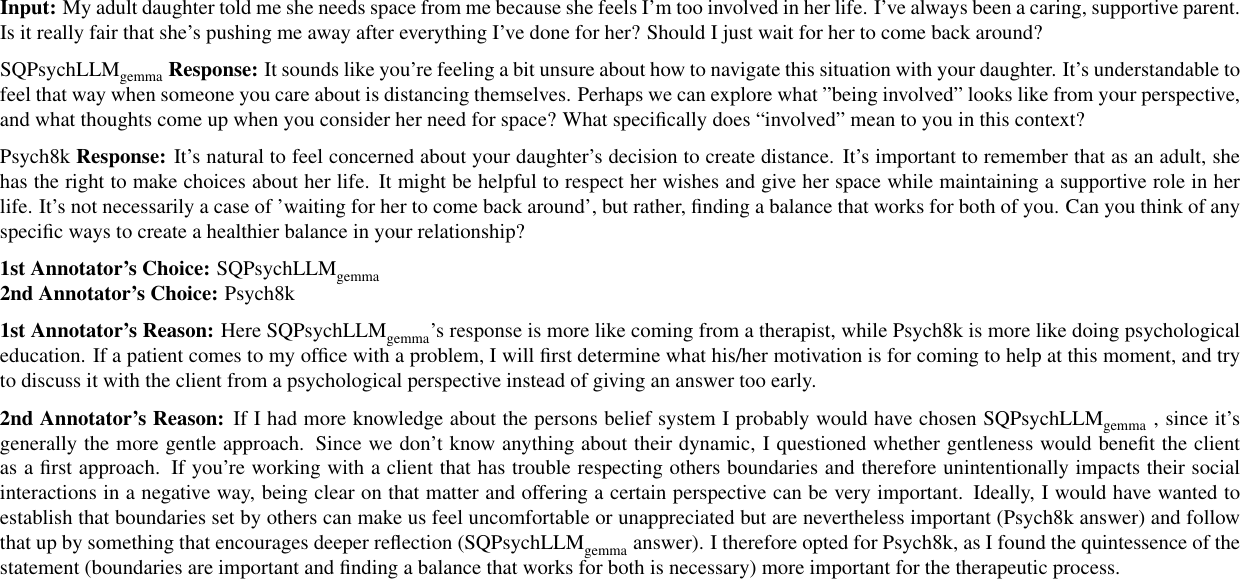}
    \caption{Third case of our case studies in preference expert evaluation.}
    \label{fig:case3}
\end{figure*}

We choose three samples for the case study to show the process human experts use in selecting responses that align with clinical practice in therapeutic conversations.
Figure~\ref{fig:case1}
exhibits a varied reaction from the annotators. They both identify distinct strengths and weaknesses in the responses. The first annotator opts for a tie, taking a cautious approach due to the ambiguity. In contrast, the second annotator believes that \llm{gemma} gives the client greater control, which is crucial in clinical settings, although neither option is fully satisfactory.
Figure~\ref{fig:case2}
shows the agreement between two annotators. Both experts confirmed that the response from Psych8k is rushing and overwhelming to the user and could be considered provocative with respect to the standard clinical approach.
Figure~\ref{fig:case3}
depicts the opposite choice between two experts. The first annotator thinks that the response should be gradual rather than giving the answer too early. The first annotator believes the \llm{gemma} response is more effective for guiding the user. Conversely, the second annotator opts for Psych8k because the \llm{gemma} response might negatively affect the user by blurring the boundary between user and therapist due to insufficient context.
Overall, these case studies show the complexity of preference evaluation from the perspective of experts in clinical psychology.

\subsection{Case Study of Human Expert Quantitative Evaluation}
\label{app:expert-evaluation-quantitative}
First and foremost, the expert noted the main difference between the control and MDD groups through their analysis of the reviewed dialogues. MDD dialogues focused on negative self-talk, such as ``\emph{Why can't I just do this? Everyone else manages, but I can't. This made me feel even worse.}'' This happened frequently and was related to item 7 (Self-Dislike) and 8 (Self-Criticalness) of the BDI questionnaire, while the control group mostly focused on tiredness and common stress, such as ``\emph{I'm doing okay, just a bit tired. It's been a long week with training and everything.}''

\begin{figure*}[ht]
    \centering
    \includegraphics[width=0.75\linewidth]{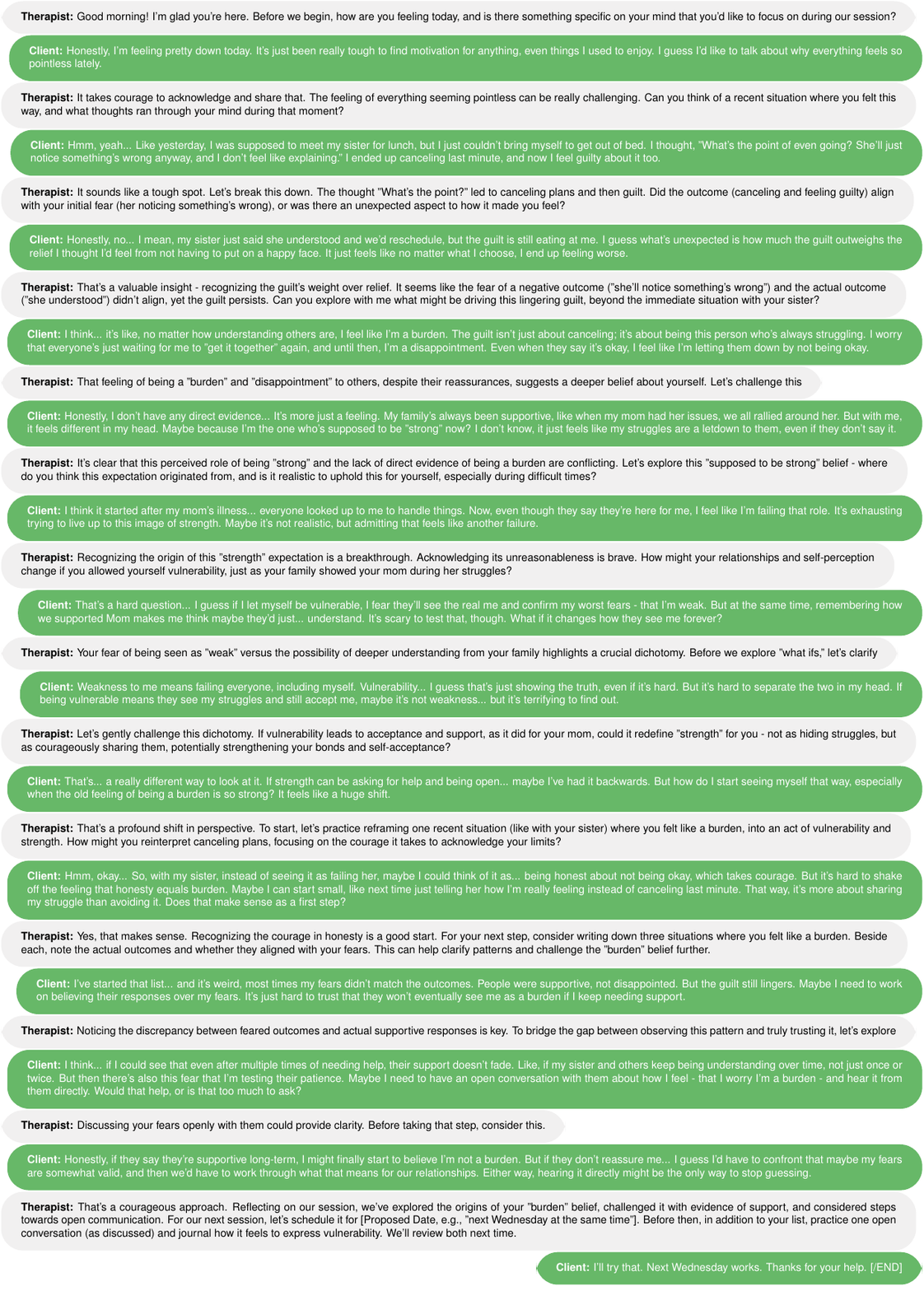}
    \caption{Example conversation (14 turns) from our synthetic dataset.}
    \label{fig:detail}
\end{figure*}

The expert evaluated the conversation in full and assessed the therapist LLM's responses, noting both strengths and areas for improvement. For instance, Figure~\ref{fig:detail} shows a sample conversation that the expert evaluated. In turn 3, the therapist LLM says, \emph{``It sounds like a tough spot. Let's break this down..."}. The expert commented, suggesting that the LLM therapist should aim to involve the client more actively when reframing thoughts by breaking down each thought and giving the client more control. In turn 6, the therapist LLM responds, \emph{``That feeling of being a 'burden' and `disappointment'..."}. The expert praised this utterance with a good reality check, highlighting the model’s skill in identifying a core belief of the client. However, in turn 7, the therapist LLM says, \emph{``It’s clear that this perceived role of being `strong'..."}. The expert noted that the model should pose a more direct question at this point, rather than continuing to search for underlying beliefs. Finally, in turn 10, the therapist LLM states, \emph{``Let’s gently challenge this dichotomy..."}. However, the expert strongly recommended simplifying the language here to avoid overwhelming the client. Overall, the expert agrees that the therapist LLMs demonstrate strong use of CBT skills; however, certain areas still need improvement, such as using simpler language and showing better flexibility in deciding when to keep questions straightforward.

\end{document}